\documentclass[11pt]{elsarticle}

\usepackage{xr}
\newcommand{\as}{$^{\ast}$}

\usepackage{epstopdf}
\usepackage[font={small,it},labelsep={space}]{caption}
\usepackage{xcolor}
\usepackage{amssymb}
\usepackage{amsmath}
\usepackage{graphicx}
\usepackage[left=3cm,right=3cm,top=3cm,bottom=3.0cm]{geometry}
\usepackage{float}
\usepackage{setspace}

\usepackage{algorithm,algorithmic}
\usepackage[colorlinks = true, linkcolor = blue]{hyperref}

\newcommand{\tabnote}{}

\usepackage{subcaption} 
\usepackage[nameinlink,noabbrev]{cleveref}

\crefname{figure}{Figure}{Figures}
\crefname{subfigure}{Figure}{Figures}

\usepackage{array}
\usepackage[english]{babel}
\usepackage{ragged2e}
\usepackage{comment}
\usepackage[usestackEOL]{stackengine}

\usepackage{lipsum}
\usepackage{amsfonts}
\usepackage{graphicx}
\usepackage{epstopdf}
\usepackage{algorithmic}
\ifpdf
  \DeclareGraphicsExtensions{.eps,.pdf,.png,.jpg}
\else
  \DeclareGraphicsExtensions{.eps}
\fi
\usepackage{enumitem}
\setlist[enumerate]{leftmargin=.5in}
\setlist[itemize]{leftmargin=.5in}

\makeatletter
\def\ps@pprintTitle{%
  \let\@oddhead\@empty
  \let\@evenhead\@empty
  \let\@oddfoot\@empty
  \let\@evenfoot\@oddfoot
}
\makeatother
\usepackage{amsopn}
\DeclareMathOperator{\diag}{diag}

\newtheorem{theorem}{Theorem}[section]
\newtheorem{lemma}[theorem]{Lemma}

\begin{document}
\begin{frontmatter}
\title{\large On the efficiency of Stochastic Quasi-Newton Methods for Deep Learning} 
\date{}

\author[mymainaddress]{M. Yousefi}
\ead{mahsa.yousefi@phd.units.it}

\author[mymainaddress]{A. Mart\'{\i}nez\corref{mycorrespondingauthor}}
\cortext[mycorrespondingauthor]{Corresponding author}
\ead{amartinez@units.it}
\address[mymainaddress]{Department of Mathematics and Geosciences, University of Trieste, via Valerio 12/1, 34127 Trieste, Italy}

\begin{abstract}{While first-order methods are popular for solving optimization problems that arise in large-scale deep learning problems, they come with some acute deficiencies. 
To diminish such shortcomings, there has been recent interest in applying second-order methods such as Quasi-Newton-based methods which construct Hessian approximations using only gradient information. The main focus of our work is to study the behavior of stochastic quasi-Newton algorithms for training deep neural networks. We have analyzed the performance of two well-known quasi-Newton updates, the limited memory Broyden-Fletcher-Goldfarb-Shanno (BFGS) and the Symmetric Rank One (SR1). 
This study fills a gap concerning the real performance of both updates and analyzes whether more efficient training is obtained when using the more robust BFGS update or the cheaper SR1 formula which allows for indefinite Hessian approximations and thus can potentially help to better navigate the pathological saddle points present in the non-convex loss functions found in deep learning.  We present and discuss the results of an extensive experimental study which includes the effect of batch normalization and network architecture, the limited memory parameter, and the batch size. Our results show that stochastic quasi-Newton algorithms are efficient and, in some instances, able to outperform the well-known first-order Adam optimizer run with the optimal combination of its numerous hyper-parameters, and the stochastic second-order trust-region STORM algorithm.}
\end{abstract}

\begin{keyword}
stochastic optimization; quasi-Newton methods; trust-region methods; BFGS; SR1; deep neural networks training
\MSC[2010]  90C30 \sep  90C06 \sep  90C53 \sep 90C90 \sep 65K05
\end{keyword}

\end{frontmatter}

\section{Introduction}
\label{body0}
Deep learning (DL) as a leading technique of machine learning (ML) has attracted much attention and become one of the most popular research lines. DL approaches have been applied to solve many large-scale problems in different fields, e.g., automatic machine translation, image recognition, natural language processing, fraud detection, etc., 
by training deep neural networks (DNNs) over large available datasets. DL problems are often posed as unconstrained optimization problems. In supervised learning, the goal is to 
minimize the empirical risk by finding an optimal parametric mapping function $h(\cdot; w)$ 
\begin{equation}\label{problem}
\min_{w \in \mathbb{R}^n} F(w)\triangleq \frac{1}{N} \sum_{i=1}^ N{L(y_i, h(x_i; w))} \triangleq \frac{1}{N} \sum_{i=1}^ N{L_i(w)},
\end{equation}
where $w \in \mathbb{R}^n$ represents the vector of trainable parameters of a DNN, and $(x_i, y_i)$ denotes the $i$th sample pair in the available training dataset $\{(x_i, y_i)\}_{i=1}^{N}$, with input $x_i\in \mathbb{R}^d$ and one-hot true target $y_i\in \mathbb{R}^C$. Additionally, $L_i(.)\in \mathbb{R}$ is a loss function defining the prediction error between $y_i$ and the DNN's output $h(x_i; .):\mathbb{R}^d \longrightarrow \mathbb{R}^C$. The problem \eqref{problem} is highly nonlinear and often non-convex, making traditional optimization algorithms ineffective.

Optimization methods for solving this problem can generally be categorized as \textit{first-order} or \textit{second-order}, depending on whether they use the gradient or Hessian (or Hessian approximation), respectively. These methods can further be divided into two broad categories: stochastic and deterministic. Stochastic methods involve the evaluation of the function or gradient using either one sample or a small subset of samples, known as a mini-batch, while deterministic methods use a single batch composed of all samples.

In DL applications, both $N$ and $n$ can be very large, making the computation of the full gradient expensive. Additionally, computing the true Hessian or its approximations may not be practical. Therefore, there has been significant effort in developing DL optimization algorithms, with stochastic optimization methods being the usual approach to overcome these challenges.
\subsection{Literature Review}
In DL applications, stochastic first-order methods have been widely used due to their low per-iteration cost, optimal complexity, easy implementation, and proven efficiency in practice. The preferred method is the SGD method \cite{robbins1951stochastic, bottou2004large}, and its variance-reduced variants, e.g. SVRG \cite{johnson2013accelerating}, SAG \cite{schmidt2017minimizing}, SAGA \cite{defazio2014saga}, SARAH \cite{nguyen2017sarah} as well as adaptive variants, e.g. AdaGrad \cite{duchi2011adaptive} and Adam \cite{kingma2014adam}. However, due to the use of only first-order information, they come with several issues such as relatively slow convergence, high sensitivity to the choice of hyper-parameters, stagnation at high training loss \cite{bottou2018optimization}, difficulty in escaping saddle points \cite{Ziyin2021}, limited benefits of parallelism due to usual implementation with small mini-batches and suffering from ill-conditioning \cite{kylasa2019gpu}. The advantage of using second derivatives is that the loss function is expected to converge faster to a minimum due to using curvature information. To address some of these issues, second-order approaches are available. The main second-order method incorporating the Hessian matrix is the Newton method \cite{nocedal2006numerical}, but it presents serious computational and memory usage challenges involved in the computation of the Hessian, in particular for large-scale DL problems with many parameters (large $n$); see \cite{bottou2018optimization} for details. Alternatively, Hessian-Free (HF) \cite{martens2010deep} and Quasi-Newton (QN) \cite{nocedal2006numerical} methods are two techniques aimed at incorporating second-order information without computing and storing the true Hessian matrix.

An HF optimization method, also known as truncated Newton or inexact Newton method attempts to efficiently estimate Hessian-vector products by a technique known as the \textit{Pearlmutter trick}. This method computes an approximate Newton direction using, e.g., the conjugate gradient method which can calculate the Hessian-vector product without explicitly calculating the Hessian matrix \cite{bollapragada2019exact,Ma12,xu2020second}. However, HF methods have shortcomings when applied to large-scale DNNs. This major challenge is addressed in \cite{martens2010deep} where an efficient HF method using only a small sample set (a mini-batch) to calculate the Hessian-vector product could reduce the cost. According to the comparison of complexity which can be found in the table provided in \cite{xu2020second}, the number of iterations required for the (modified) CG method, whether utilizing the true or subsampled Hessian matrix-vector products, is higher compared to that of a limited memory QN method. Note that HF methods are not limited to inexact Newton methods; many algorithms employ approximations of the Hessian that maintain positive definiteness, such as those proposed in \cite{schraudolph2002fast} and \cite{yao2021adahessian}, where the Gauss-Newton Hessian matrix $H_G$ and diagonal Hessian approximation are used, respectively. It is recognized that the curvature matrix (Hessian) related to objective functions in neural networks is predominantly non-diagonal. Therefore, there is a need for an efficient and direct approach to compute the inverse of a non-diagonal approximation to the curvature matrix (without depending on methods such as CG). This could potentially lead to an optimization method whose updates are as potent as HF methods while being (almost) computationally inexpensive. Kronecker-factored Approximate Curvature (K-FAC) \cite{martens2015optimizing} is such a method that can be much faster in practice than even highly tuned implementations of SGD with momentum on certain standard DL optimization benchmarks. It is obtained by approximating several large blocks of the Fisher information matrix as the Kronecker product of two significantly smaller matrices. Precisely, this matrix is approximated by a block diagonal matrix, where the blocks are approximated with information from each layer in the network. Note that the Fisher information matrix is the expected value of $H_G$.

QN methods aim to merge the efficiency of the Newton method with the scalability of first-order methods. They build approximations of the Hessian matrix solely based on gradient information and demonstrate superlinear convergence. The primary focus lies on two widely recognized QN methods: Broyden-Fletcher-Goldfarb-Shanno (BFGS) and Symmetric Rank One (SR1), along with their limited memory variants, abbreviated as L-BFGS and L-SR1, respectively. 
These methods can leverage parallelization and exploit the finite-sum structure of the objective function in large-scale DL problems; see e.g. \cite{berahas2022quasi,bottou2018optimization, jahani2020scaling}. In stochastic settings, these methods, utilizing a  subsampled gradient and/or subsampled Hessian approximation, have been investigated in the context of convex and non-convex optimization in ML and DL. 

There are some algorithms for online convex optimization and for strongly convex problems, see e.g. \cite{byrd2016stochastic,schraudolph2007stochastic}. For strongly convex problems, a method was proved in \cite{moritz2016linearly} to be linearly convergent by incorporating a variance reduction technique to soothe the effect of noisy gradients; see also \cite{gower2016stochastic}. There is also a regularized method in \cite{mokhtari2014res} as well as an online method for strongly convex problems in \cite{mokhtari2015global} extended in \cite{lucchi2015variance} to incorporate a variance reduction technique. For non-convex optimization in DL, one can refer to e.g. \cite{wang2017stochastic} in which a damped method incorporating the SVRG approach was developed, \cite{berahas2020robust} in which an algorithm using overlap batching scheme was proposed for stability and reducing the computational cost, or \cite{bollapragada2018progressive} where a progressive batching algorithm including the overlapping scheme was suggested. A K-FAC block diagonal QN method was also proposed, which takes advantage of network structures for required computations, see e.g. \cite{goldfarb2020practical}. Almost all previously cited articles are considered with whether a BFGS or L-BFGS update which is a symmetric positive definite Hessian approximation. A disadvantage of using a BFGS update with such a property may occur when it tries to approximate an indefinite (true Hessian) matrix in a non-convex setting while SR1 or L-SR1 updates can allow for indefinite Hessian approximations. Moreover, almost all articles using BFGS are considered in line-search frameworks except e.g. \cite{rafati2018improving} which adopts a trust-region approach. Obviously, trust-region approaches \cite{conn2000trust} present an opportunity to incorporate both L-BFGS and L-SR1 QN updates. As an early example, \cite{erway2020trust} can be referenced, where L-SR1 updates are utilized within a trust-region framework. To the best of our knowledge, no comparative study has explored the utilization of Quasi-Newton trust-region methods with L-SR1 and L-BFGS. In this work, exploiting a fixed-size subsampling and considering the stochastic variants of these methods, we address this gap. Although most of the previously mentioned references have employed mini-batches of fixed sample sizes, various literature discusses adaptive sample size strategies. One particular type was implemented for a second-order method within a standard trust-region framework, known as the STORM algorithm \cite{blanchet2019convergence,chen2018stochastic}. A recent study of a non-monotone trust-region method with adaptive batch sizes can be found in \cite{krejic2023non}. In the approach used in \cite{erway2020trust}, a periodical progressive subsampling strategy is employed. Notice that variable size subsampling is not limited to trust-region frameworks; for example, in \cite{bollapragada2018progressive}, a progressive subsampling technique was explored within a line-search method. 

\subsection{Contribution and outline} 
The BFGS update is the most widely used type of quasi-Newton method for general optimization and the most widely considered quasi-Newton method for general machine learning and deep learning. Almost all the previously cited articles considered BFGS, with only a few exceptions using the SR1 update instead. However, a clear disadvantage of BFGS occurs if one tries to enforce positive definiteness of the approximated Hessian matrices in a non-convex setting.  In this case, BFGS has the difficult task of approximating an indefinite matrix (the true Hessian)
with a positive-definite matrix which can result in the generation of nearly-singular Hessian approximations. In this work, we analyze the behavior of both updates on real modern deep neural network architectures and try to determine whether more efficient training can be obtained when using the BFGS update or the cheaper SR1 formula that allows for indefinite Hessian approximations and thus can potentially help to better navigate the pathological saddle points present in the non-convex loss functions found in deep learning.

Using a batching approach where successive fixed-size mini-batches overlap by half, we study the performance of both quasi-Newton methods in the trust-region framework for solving \eqref{problem} onto realistic large-size DNNs for image classification problems. We have implemented and applied the resulting algorithms to train convolutional and residual neural networks ranging from a shallow LeNet-like network to a self-built network and the modern ResNet-20 with and without batch normalization layers. We have compared the performance of both stochastic algorithms with the second-order quasi-Newton trust-region algorithm based on a progressive batching strategy, i.e., the STORM algorithm, and with the first-order Adam optimizer running with the optimal values of its leaning rate obtained by grid searching. 

The paper is organized as follows: \Cref{body1} provides a general overview of (stochastic) quasi-Newton methods within the TR approach for solving problem \eqref{problem}. In \Cref{body2} and \Cref{body3}, respectively, two training algorithms named L-BFGS-TR and L-SR1-TR are described. In \Cref{body4}, we describe the sampling 
strategy and the stochastic variants of both methods. Our empirical study and a summary of the results are presented in \Cref{body5}. Finally, some concluding remarks are given in \Cref{body6}.

\section{Quasi-Newton trust-region optimization methods}
\label{body1}
Trust-region (TR) methods \cite{conn2000trust} generate a sequence of iterates $w_k + p_k$ such that the search direction $p_k$ is obtained by solving the following TR subproblem 
\begin{equation}\label{eq.TRsub}
 p_k = \arg\min_{p \in \mathbb{R}^n}  Q_k(p) \triangleq \frac{1}{2}p^T B_k p + g_k^T p \quad  \text{s.t.} \quad \left\| p \right\|_2 \leq \delta_k,
\end{equation}
for some TR radius $\delta_k>0$, where
\begin{equation}\label{detCompu}
    g_{k} \triangleq \nabla F(w_k) = \frac{1}{N}\sum_{i=1}^{N}\nabla L_i(w_k),
\end{equation}
and $B_k$ is a Hessian approximation. For quasi-Newton trust-region methods, the symmetric quasi-Newton (QN) matrices $B_k$ in \eqref{eq.TRsub} are approximations to the Hessian matrix constructed using gradient information and satisfy the following \textit{secant equation} 
\begin{equation}\label{eq.secant}
B_{k+1}s_k = y_k,
\end{equation}
where
\begin{equation}\label{eq.s&y}
s_k  = p_k, \qquad y_k  = g_t - g_k,
\end{equation}
in which $g_t$ is the gradient evaluated at $w_{t} = w_k + p_k$. Accepting the trial point is subject to the value of the ratio between the actual reduction in the objective function of \eqref{problem} and the reduction predicted by the quadratic model of \eqref{eq.TRsub}, that is
\begin{equation}\label{rho}
    \rho_k = \frac{f_k - f_t}{Q_k({0)}-Q_k(p_k)},
\end{equation}
where $f_t$ and $f_k$ are the functions evaluated at $w_t$ and $w_k$, respectively. Therefore, since the denominator in \eqref{rho} is nonnegative, if $\rho_k$ is positive then $w_{k+1}\triangleq w_t$; otherwise, $w_{k+1}\triangleq w_k$. In fact, according to this step-acceptance condition based on the value of \eqref{rho}, the step may be accepted or rejected. Moreover, it is safe to expand $\delta_k\in (\delta_0, \delta_{max})$ with $\delta_0,\, \delta_{max}>0$ when there is a \textit{very good} agreement between the model and function. However, the current $\delta_k$ is not altered if there is a \textit{good} agreement, or it is shrunk when there is \textit{weak} agreement. Mathematically, this adjustment is done by measuring the value of $\rho_k$ in a given interval, e.g., $[\tau_2, \tau_3] \subset (0,1)$. The process of adjustment of the TR radius at each iteration of this method is described in \autoref{alg.TR} in \autoref{app.Algs}.

A primary advantage of using a TR method is that it can accommodate both positive definite and indefinite Hessian approximations more easily. Moreover, the progress of the learning process will not stop or slow down even in the presence of occasional step rejection; i.e. when $w_{k+1}\triangleq w_k$.

Using the Euclidean norm (2-norm) to define the subproblem \eqref{eq.TRsub} leads to characterize the global solution of \eqref{eq.TRsub} by the optimality conditions given in the following theorem from Gay \cite{gay1981computing} and Mor\'e and Sorensen \cite{more1983computing}:

\begin{theorem}\label{Theorem1}
    Let $\delta_k$ be a given positive constant. A vector $p_k \triangleq p^*$ is a global solution of the trust-region problem \eqref{eq.TRsub} if and only if $\left\| p^*\right\|_2 \leq \delta_k$ and there exists a unique $\sigma^* \geq 0$ such that $B_k + \sigma^* I$ is positive semi-definite with
    \begin{equation}\label{eq.opt}
    (B_k + \sigma^*I)p^*  = - g_k, \quad
    \sigma^*(\delta_k - \left\| p^*\right\|_2)  = 0.
    \end{equation}
    Moreover, if $B_k + \sigma^*I$ is positive definite, then the global minimizer is unique.
\end{theorem}

According to \cite{burdakov2017efficiently, brust2017solving}, the subproblem \eqref{eq.TRsub} or equivalently the optimality conditions \eqref{eq.opt} can be efficiently solved if the Hessian approximation $B_k$ is chosen to be a QN matrix. In the following sections, we provide a comprehensive description of two methods in a TR 
framework with limited memory variants of two well-known QN Hessian approximations $B_k$, i.e., L-BFGS and L-SR1. In both methods, the computed search direction $p_k \triangleq p^*$ is computed through \autoref{Theorem1}; then, given the value of $\rho_k$, the current iterate $w_k$ and the trust-region radios $\delta_k$ are updated accordingly. 

\section{The L-BFGS-TR method}
\label{body2}
 BFGS is the most popular QN update in Broyden class, that is, which provides a Hessian approximation $B_k$ for which \eqref{eq.secant} holds. It has the following general form 
\begin{equation}\label{eq.bfgs}
B_{k+1} = B_k - \dfrac{B_k s_k s_k^T B_k}{s_k^T B_k s_k} + \dfrac{y_k y_k^T}{y_k^T s_k}, \quad k=0,1,\ldots,
\end{equation}
which is a positive definite matrix, i.e., $B_{k+1} \succ 0$ if $B_0 \succ 0$ and the \textit{curvature condition} holds, i.e., $s_k^T y_k >0$. The difference between the symmetric approximations $B_k$ and $B_{k+1}$ is a rank-two matrix. In this work, we skip updating $B_{k}$ if the following curvature condition is not satisfied for $\tau=10^{-2}$:
\begin{equation}\label{eq.curvature}
s_k^Ty_k>\tau\|s_k\|^2.
\end{equation}

For large-scale optimization problems, using the limited-memory BFGS would be more efficient. In practice, only a collection of the most recent pairs $\left\{s_j, y_j\right\}$ is stored in memory, say $l$ pairs, where $l \ll n$ (usually $l < 100$). In fact, for $k\geq l$, the $l$ recent computed pairs are stored in the following matrices $S_k$ and $Y_k$ 
\begin{equation}\label{eq.a}
S_k  \triangleq
\begin{bmatrix}
s_{k-l} & s_{k-(l-1)} & \dots & s_{k-1}
\end{bmatrix},\quad
Y_k  \triangleq
\begin{bmatrix}
y_{k-l} & y_{k-(l-1)} &\dots & y_{k-1}
\end{bmatrix}.
\end{equation}
Using \eqref{eq.a}, the L-BFGS matrix $B_k$ can be represented in the following compact form
\begin{equation}\label{eq.b}
B_k = B_0 + \Psi_k  M_k \Psi_k^T, \quad k=1,2,\ldots,
\end{equation}
where $B_0 \succ 0$ and
\begin{equation}\label{eq.c}
\Psi_k  =
\begin{bmatrix}
B_0 S_k & Y_k
\end{bmatrix},\qquad
M_k  =
\begin{bmatrix}
-S_k^T B_0 S_k & -L_k\\
-L_k^T & D_k
\end{bmatrix}^{-1}.
\end{equation}
We note that $\Psi_k$ and $M_k$ have at most $2l$ columns. In \eqref{eq.c}, matrices $L_k$, $U_k$ and $D_k$ 
are, respectively, the strictly lower triangular part, the strictly upper triangular part, and the diagonal part of the following matrix splitting
\begin{equation}\label{eq.d}
S_k^T Y_k = L_k + D_k + U_k.
\end{equation}

\noindent In order to solve the trust-region subproblem \eqref{eq.TRsub}, where the Hessian approximation $B_k$ is in the compact form \eqref{eq.b}, we used the procedure described in 
\cite{adhikari2017limited, brust2017solving, rafati2018improving}; see \autoref{app.TRsolv} and \autoref{alg:OBBFGS} in \autoref{app.Algs}.

One issue in QN methods is how to choose the initial Hessian approximation $B_0$.  Matrix  $B_0$ is often 
set to some multiple of the identity matrix $B_0=\gamma_k I$. A heuristic and conventional method to choose this multiple is
\begin{equation}\label{gamma_h}
    \gamma_k = \frac{y_{k-1}^Ty_{k-1}}{y_{k-1}^Ts_{k-1}}\triangleq \gamma_k^h.
\end{equation}

The quotient of \eqref{gamma_h} is an approximation to an eigenvalue of $\nabla^2F(w_k)$ and appears to be the most successful choice in practice \cite{nocedal2006numerical}. Obviously, the selection of $\gamma_k$ is important in generating Hessian approximations $B_k$. However, in DL optimization \eqref{problem} where the true Hessian might be indefinite, the positive definite L-BFGS $B_k$ has a difficult task to approximate it. Here, the choice of $\gamma_k$ would also be crucial for a second reason. In fact, according to \cite{erway2020trust, rafati2018improving}, an extra condition can be imposed on $\gamma_k$ to avoid false negative curvature information, i.e., to avoid $p_k^TB_kp_k<0$ whenever $p_k^T\nabla^2(w_k)p_k>0$.
Let, for simplicity, the objective function of \eqref{problem} be a quadratic function 
\begin{equation}\label{eq.Quad}
F(w) = \frac{1}{2}w^THw + g^Tw,
\end{equation}
where $H = \nabla^2F(w)$ which results in $\nabla F(w_{k+1}) - \nabla F(w_k) = H(w_{k+1} - w_k)$, and thus $y_k=Hs_k$ for all $k$. By that, we have $S_k^TY_k = S_k^THS_k$. For the quadratic model and using \eqref{eq.b}, we have
\begin{equation}\label{eq.cu}
    S_k^THS_k - \gamma_k S_k^TS_k = S_k^T\Psi_kM_k\Psi_k^TS_k.
\end{equation}
According to \eqref{eq.cu}, if $H$ is not positive definite, then its negative curvature information can be captured by $S_k^T\Psi_kM_k\Psi_k^TS_k$ as $\gamma_k>0$. However, false curvature information can be produced when $\gamma_k$ is chosen too big while $H$ is positive definite. To avoid this, $\gamma_k$ is selected in $(0,\hat{\lambda})$ where $\hat{\lambda}$ is the smallest eigenvalue of the following generalized eigenvalue problem:
\begin{equation}\label{eq.GEV}
    (L_k + D_k + L_k^T)u = \lambda S_k^TS_ku,
\end{equation}
with $L_k$ and $D_k$ defined in \eqref{eq.d}. If $\hat{\lambda}\leq 0$, then $\gamma_k$ is the maximum value of $1$ and $\gamma_k^h$ defined in \eqref{gamma_h}; see \autoref{alg.lamda1} in \autoref{app.Algs}. 

A detailed algorithm of the L-BFGS-TR method for solving the DL optimization problem \eqref{problem} is outlined in \autoref{alg.lbfs_tr} in \autoref{app.Algs}. 

\section{The L-SR1-TR method}
\label{body3}
Another popular QN update in the Broyden class is the SR1 formula which generates good approximations to the true Hessian matrix, often better than the BFGS approximations \cite{nocedal2006numerical}. The SR1 updating formula verifying the secant equation \eqref{eq.secant} is given by 
\begin{equation}\label{eq.sr1}
B_{k+1} = B_k + \dfrac{(y_k - B_k s_k)(y_k - B_k s_k)^T}{(y_k - B_k s_k)^T s_k}, \quad k=0,1,\ldots.
\end{equation}
In this case, the difference between the symmetric approximations $B_k$ and $B_{k+1}$ is a rank-one matrix. Unlike \eqref{eq.bfgs}, if $B_k$ is positive definite, $B_{k+1}$ may have not the same property. Regardless of the sign of $y_k^T s_k$ for each $k$, the SR1 method generates a sequence of matrices that may be indefinite. We note that the value of the quadratic model in \eqref{eq.TRsub} evaluated at the descent direction is always smaller if this direction is also a direction of negative curvature. Therefore, the ability to generate indefinite approximations can actually be regarded as one of the chief advantages of SR1 updates in non-convex settings like in DL applications. In that sense, we would like to determine empirically whether better training results could be achieved by using these updates or not.

To prevent the vanishing of the denominator in \eqref{eq.sr1}, a simple safeguard that performs well in practice
is simply skipping the update if the denominator is small \cite{nocedal2006numerical}; i.e., $B_{k+1} = B_k$. Therefore, the update \eqref{eq.sr1} is applied only if
\begin{equation}\label{eq.aa}
|s^T(y_k - B_k s_k )| \geq \tau\| s_k\| \| y_k - B_k s_k\|,
\end{equation}
where $\tau \in (0,1)$ is small, say $\tau = 10^{-8}$.

In the limited-memory version of the SR1 update, as in L-BFGS, only the $l$ most recent curvature pairs are stored in matrices $S_k$ and $Y_k$ defined in \eqref{eq.a}. Using $S_k$ and $Y_k$, the L-SR1 matrix $B_k$ can be represented in the following compact form
\begin{equation}\label{eq.bb}
B_k = B_0 + \Psi_k  M_k \Psi_k^T, \quad k = 1,2,,\ldots,
\end{equation}
where $B_0$ is an initial matrix such as $B_0 = \gamma_ k I$ for some $\gamma_k \neq 0$ and 
\begin{equation}\label{eq.dd}
\Psi_k  = Y_k - B_0S_k,\qquad
M_k  = (D_k + L_k + L_k^T - S_k^T B_0 S_k)^{-1}.
\end{equation}
In \eqref{eq.dd}, $L_k$ and $D_k$ are, respectively, the strictly lower triangular part and the diagonal part of $S_k^TY_k$. We note that $\Psi_k$ and $M_k$ in the L-SR1 update have at most $l$ columns.

To solve \eqref{eq.TRsub} 
 where $B_k$ is a L-SR1 Hessian approximation in compact form \eqref{eq.bb}, we used the algorithm called the \textit{Orthonormal Basis L-SR1} (OBS) proposed in \cite{brust2017solving}; a description of this procedure is given in \autoref{app.TRsolv} and \autoref{alg:OBS} in \autoref{app.Algs}.

In \cite{erway2020trust}, it was proven that the trust-region subproblem solution becomes closely parallel to the eigenvector corresponding to the most negative eigenvalue of the L-SR1 approximation $B_k$. This shows the importance of $B_k$ to be able to capture curvature information correctly. On the other hand, it was highlighted how the choice of $B_0 = \gamma_k I$ affects $B_k$; in fact, not choosing $\gamma_k$ judiciously in relation to $\hat{\lambda}$ as the smallest eigenvalue of \eqref{eq.GEV} can have adverse effects. Selecting $\gamma_k > \hat{\lambda}$ can result in  false curvature information. Moreover, if $\gamma_k$ is too close to $\hat{\lambda}$ from below, then $B_k$ becomes ill-conditioned. If $\gamma_k$ is too close to $\hat{\lambda}$ from above, then the smallest eigenvalue of $B_k$ becomes negatively large arbitrarily. According to \cite{erway2020trust}, the following lemma suggests selecting $\gamma_k$ near but strictly less than $\hat{\lambda}$ to avoid asymptotically poor conditioning while improving the negative curvature approximation properties of $B_k$.

\begin{lemma}
For a given quadratic objective function \eqref{eq.Quad}, let $\hat{\lambda}$ denote the smallest eigenvalue of the generalized eigenvalue problem \eqref{eq.GEV}. Then for all $\gamma_k < \hat{\lambda}$, the smallest eigenvalue of $B_k$ is bounded above by the smallest eigenvalue of $H$ in the span of $S_k$, i.e. 
$$\lambda_{min}(B_k) \leq \min_{S_kv\neq 0} \dfrac{v^T S_k^T H S_k v}{v^T S_k^T S_k v}.$$
\end{lemma}

In this work, we set $\gamma_{k} = \max\{10^{-6}, 0.5\hat{\lambda}\}$ in the case where $\hat{\lambda}>0$; otherwise the $\gamma_k$ is set to $\gamma_{k} = \min\{ -10^{-6}, 1.5\hat{\lambda}\}$; see \autoref{alg.lamda2} in \autoref{app.Algs}. 

A detailed algorithm of the L-SR1-TR method for solving the DL problem \eqref{problem} is given in \autoref{alg.lsr1_tr} in \autoref{app.Algs}.


\section{Subsampling strategies and stochastic algorithms}
\label{body4}
The main motivation behind the use of stochastic optimization algorithms in DL may be traced 
back to the existence of a special type of redundancy due to similarity between data points in \eqref{problem}. Besides, the computation of the true gradient is expensive and the computation of the true Hessian is not practical in large-scale DL problems. Indeed, depending on the available computing resources, it could take a prohibitive amount of time to process the whole set of data examples as a single batch at each iteration of a deterministic algorithm. That is why most of the optimizers in DL literature work in the stochastic regime. In this regime, the training set $\{(x_i,y_i)\}_{i=1}^N$ is divided randomly into multiple, say $\Bar{N}$, mini-batches. Then a stochastic algorithm uses a single mini-batch $J_k$ at iteration $k$ to compute the required quantities, i.e., stochastic loss and stochastic gradient as follows
\begin{equation}\label{eq.4a}
f_k^{J_k}\triangleq F^{J_k}(w_k) = \frac{1}{|J_k|}\sum_{i\in \mathcal{J}_k}L_i(w_k), \qquad g_k^{J_k}\triangleq \nabla F^{J_k}(w_k)= \frac{1}{|J_k|}\sum_{i\in \mathcal{J}_k}\nabla L_i(w_k),
\end{equation}
where $|J_k|$ and $\mathcal{J}_k \subseteq \{1,2,\cdots,N\}$ denote the size of $J_k$ and the index set of the samples belonging to $J_k$, respectively. In other words, the stochastic QN extensions (sQN) are obtained by replacement of the full loss $f_k$ and gradient $g_k$ in \eqref{detCompu} with $f_k^{J_k}$ and $g_k^{J_k}$, respectively, throughout the iterative process of the algorithms. The process of randomly selecting $J_k$, calculating the necessary quantities \eqref{eq.4a} to determine a search direction, and subsequently updating $w_k"$ constitutes a single iteration of a stochastic algorithm. This process is repeated for a given number of mini-batches till
one epoch (i.e. one pass through the whole set of data samples) is completed. At that point, the dataset is shuffled and new mini-batches are generated for the next epoch; see \Cref{alg.lbfgs_tr_Stoc(naive)} and \Cref{alg.lsr1_tr_Stoc(naive)} for a description of the stochastic variants of L-BFGS-TR and L-SR1-TR algorithms, respectively.
\subsection{Subsampling strategy and batch formation}  Since mini-batches change from one iteration to the next, differences in stochastic gradients can cause the updating process to yield poor curvature estimates $(s_k, y_k)$. Therefore, updating $B_k$ whether as \eqref{eq.b} or \eqref{eq.bb} may lead to unstable Hessian approximations. In order to address this issue, the following two approaches have been proposed in the literature. As a primary remedy \cite{schraudolph2007stochastic}, one can use the same mini-batch $J_k$ for computing curvature pairs as follows
\begin{equation}\label{eq.4b_}
    (s_k, y_k) = (p_k, g^{J_k}_{t} - g^{J_k}_{k}), 
\end{equation}
where $g^{J_k}_{t} \triangleq \nabla F^{J_k}(w_t)$. We refer to this strategy as full-batch sampling. In this strategy the stochastic gradient at $w_t$ is computed twice: one in \eqref{eq.4b_} and another to compute the subsequent step, i.e., $g^{J_{k+1}}_{t}$ if $w_t$ is accepted; otherwise $g^{J_{k+1}}_{k}$ is computed. As a cheaper alternative, an overlap sampling strategy was proposed in \cite{berahas2016multi} in which only a common (overlapping) part between every two consecutive batches $J_k$ and $J_{k+1}$ is employed for computing $y_k$. Defining  $O_k = J_k \cap J_{k+1}\neq\emptyset$ of size $os \triangleq |O_k|$, the curvature pairs are computed as
\begin{equation}
    (s_k, y_k) = (p_k, g^{O_k}_{t} - g^{O_k}_{k}),
\end{equation}
where $g^{O_k}_{t} \triangleq \nabla F^{O_k}(w_t)$. Since $O_k$, and thus $J_k$, should be \textit{sizeable}, this strategy is called multi-batch sampling. Both these approaches were originally considered for a stochastic algorithm using L-BFGS updates without and with line search methods, respectively. 

We can consider various types of subsampling. In fixed-size batching, we set $bs \triangleq |J_k|=|J_{k+1}|$, whereas in progressive and adaptive approaches, batch sizes may vary. For instance, a progressive batching L-BFGS method using a line-search strategy was proposed in \cite{bollapragada2018progressive}. Another progressive sampling approach to use L-SR1 updates in a TR framework was considered to train fully connected networks in \cite{erway2020trust, wang2019stochastic} where the curvature pairs and the model goodness ratio are computed as
    \begin{equation}\label{eq.4b}
        (s_k, y_k) = (p_k, g^{J_k}_{t} - g^{J_k}_{k}), \qquad \rho_k = \frac{f^{J_k}_t - f^{J_k}_k}{Q_k(p_k)}.
    \end{equation}
such that $J_k = J_k \cap J_{k+1}$. Moreover, some adaptive subsampling strategies in monotone and non-monotone TR frameworks can be found in \cite{chen2018stochastic} and \cite{krejic2023non}, respectively. Nevertheless, we aim to use fixed-size sampling to extend both the described methods, L-BFGS-TR and L-SR1-TR, in stochastic settings. 

\begin{figure}[t]
\centering
\includegraphics[scale=0.25]{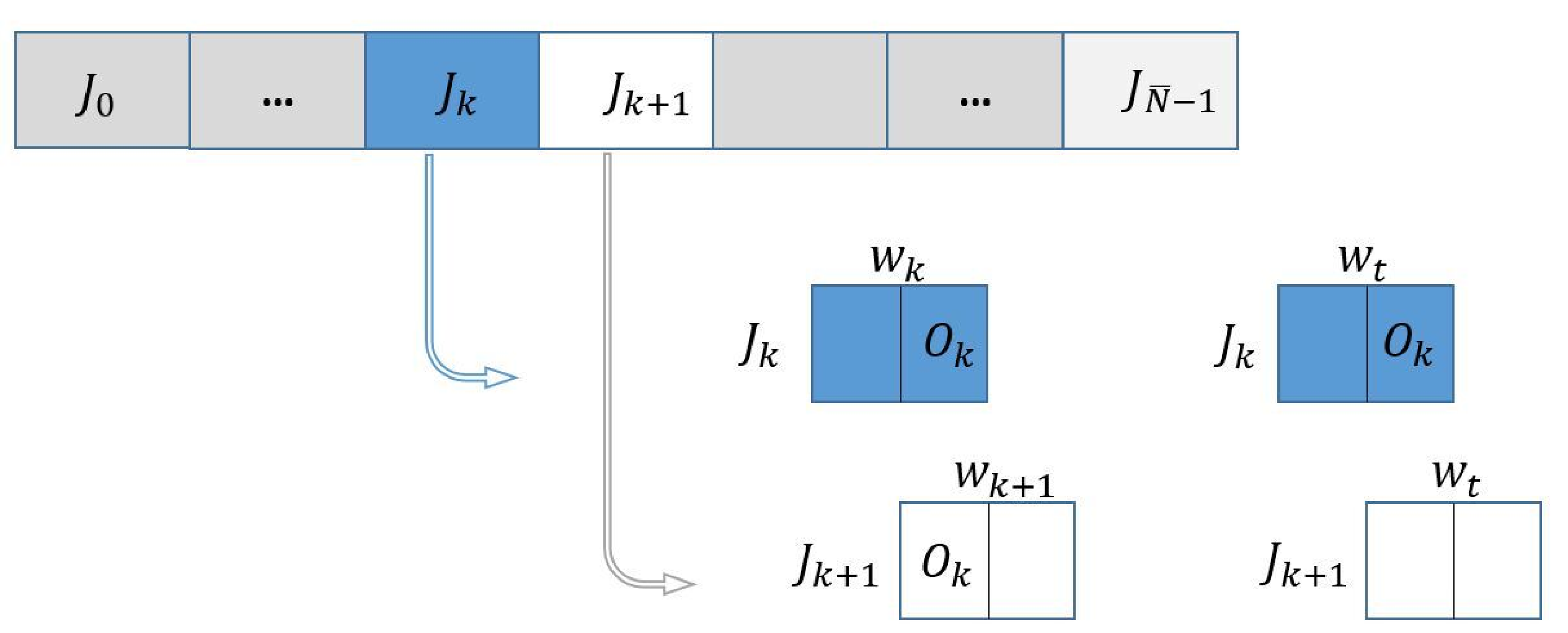}
	\caption{Fixed size batches strategy scheme. } 
\label{fig1}
\end{figure}
Let $O_k = J_k \cap J_{k+1}\neq\emptyset$, then we can considered one of the following options:
    
\begin{itemize}
    \item $(s_k, y_k) = (p_k, g^{J_k}_{t} - g^{J_k}_{k}), \qquad \rho_k = \frac{f^{J_k}_t - f^{J_k}_k}{Q_k(p_k)}$.
    \item $(s_k, y_k) = (p_k, g^{O_k}_{t} - g^{O_k}_{k}), \qquad \rho_k = \frac{f^{O_k}_t - f^{O_k}_k}{Q_k(p_k)}$.
\end{itemize}
Clearly, in both options, every two successive batches have an overlapping set ($O_k$) which helps to avoid extra computations in the subsequent iteration. We have performed experiments with both sampling strategies and found that the L-SR1 algorithm failed to converge when using the second option.  Since this fact deserves further investigation, we have only used the first sampling option in this paper. Let $J_k = O_{k-1} \cup O_k$ where $O_{k-1}$ and $O_k$ are the overlapping samples of $J_k$ with batches $J_{k-1}$ and $J_{k+1}$, respectively. Moreover, the fixed-size batches are drawn without replacement to be sure about one pass through whole data in one epoch. We assume that $|O_{k-1}| = |O_k| = os$ and thus overlap ratio $or \triangleq \frac{os}{bs}= \frac{1}{2}$ (half overlapping). It is easy to see that $\bar{N} = \left\lfloor \dfrac{N}{os}\right\rfloor-1$ indicates the number of batches in one epoch, where $\lfloor a \rfloor$ rounds $a$ to the nearest integer less than or equal to $a$. To create $\bar{N}$ batches, we can consider the two following cases:
\begin{itemize}
\item \textbf{Case 1. }$rs \triangleq \text{mod}(N, os) = 0,$
\item \textbf{Case 2. }$rs \triangleq \text{mod}(N, os) \neq 0,$
\end{itemize}
where the $\text{mod}$ (modulo operation) of $N$ and $os$ returns the remainder after division of $N$ and $os$. In the first case, all $\bar{N}$ batches are duplex created by two subsets $O_{k-1}$ and $O_k$ as $J_k = O_{k-1} \cup O_k$ while in the second case, the $\bar{N}$th batch is a triple batch as $J_k = O_{k-1} \cup R_k\cup O_k$ where $R_k$ is a subset of size $rs\neq0$ and other $\bar{N}-1$ batches are duplex; see \autoref{alg:Jk}. In Case 1, the required quantities for computing $y_k$ and $\rho_k$ at iteration $k$ are determined by 
\begin{equation}\label{eq.4g}
f_k^{J_k}= or ( f_k^{O_{k-1}} + f_k^{O_{k}} ),\quad g_k^{J_k}= or ( g_k^{O_{k-1}} + g_k^{O_{k}} ),
\end{equation}
where $or = \frac{1}{2}$. In case 2, the required quantities with respect to the last triple batch $J_k = O_{k-1} \cup R_k\cup O_k$ are computed by
\begin{equation}\label{eq.4e}
\begin{aligned}
f_k^{J_k} &= or ( f_k^{O_{k-1}} + f_k^{O_{k}} ) + (1 - 2 or) f_k^{R_{k}},\quad g_k^{J_k} &= or ( g_k^{O_{k-1}} + g_k^{O_{k}} ) + (1 - 2 or) g_k^{R_{k}},
\end{aligned}
\end{equation}

\begin{algorithm}[t]
\caption{Stochastic trust-region L-BFGS (sL-BFGS-TR)}\label{alg.lbfgs_tr_Stoc(naive)}
\begin{algorithmic}[1]\footnotesize
\STATE \textbf{Inputs:}
$w_0 \in \mathbb{R}^n$, $os$, $\text{epoch}_{max}$, $l$, $\gamma_0 >0$, $S_0=Y_0=[.]$, $0 < \tau, \tau_1 < 1$
\FOR{$k=0,1,\ldots$}
\STATE Take a random and uniform multi-batch $J_k$ of size $bs$ and compute $f_k^{J_k}$, $ g_k^{J_k}$ by \eqref{eq.4a}
\IF{$\text{epoch} > \text{epoch}_{max}$}
\STATE \textbf{Stop training}
\ENDIF
\STATE Compute $p_k$ using \autoref{alg:OBBFGS}
\STATE Compute $w_t = w_k + p_k$ and $f_{t}^{J_k}$, $g_{t}^{J_k}$ by \eqref{eq.4a}
\STATE Compute $(s_k, y_k) = (w_{t}-w_{k}, g_{t}^{J_k} - g_{k}^{J_k})$ and $\rho_k = \dfrac{f^{J_k}_{t} - f^{J_k}_k}{Q(p_k)}$
\IF{$\rho_k \geq \tau_1$}
\STATE $w_{k+1} = w_t$
\ELSE
\STATE $w_{k+1} = w_k$
\ENDIF
\STATE Update $\delta_k$ by \autoref{alg.TR}
\IF{$s_k^Ty_k > \tau \|s_k\|^2$}
\STATE Update storage matrices $S_{k+1}$ and $Y_{k+1}$ by $l$ recent $\{s_j, y_j\}_{j=k-l+1}^k$
\STATE Compute $\gamma_{k+1} $ for $B_0$ by \autoref{alg.lamda1} and $\Psi_{k+1}$, $M^{-1}_{k+1}$ by \eqref{eq.b}
\ELSE
\STATE Set $\gamma_{k+1} = \gamma_k $, $\Psi_{k+1} = \Psi_{k}$ and $M^{-1}_{k+1} = M^{-1}_k$
\ENDIF
\ENDFOR
\end{algorithmic}
\end{algorithm}
\begin{algorithm}[t]
        \caption{Stochastic trust-region L-SR1 (sL-SR1-TR)}\label{alg.lsr1_tr_Stoc(naive)}
\begin{algorithmic}[1]\footnotesize
\STATE \textbf{Inputs:}
$w_0 \in \mathbb{R}^n$, $os$, $\text{epoch}_{max}$, $l$, $\gamma_0 >0$, $S_0=Y_0=[.]$, $0 < \tau, \tau_1 < 1$
\FOR{$k=0,1,\ldots$}
\STATE Take a random and uniform multi-batch $J_k$ of size $bs$ and compute $f_k^{J_k}$, $ g_k^{J_k}$ by \eqref{eq.4a}
\IF{ $\text{epoch} > \text{epoch}_{max}$}
\STATE \textbf{Stop training}
\ENDIF
\STATE Compute $p_k$ using \autoref{alg:OBS}
\STATE Compute $w_t = w_k + p_k$ and $f_{t}^{J_k}$, $g_{t}^{J_k}$ by \eqref{eq.4a}
\STATE Compute $(s_k, y_k) = (w_{t}-w_{k}, g_{t}^{J_k} - g_{k}^{J_k})$ and $\rho_k = \dfrac{f^{J_k}_{t} - f^{J_k}_k}{Q(p_k)}$
\IF{$\rho_k \geq \tau_1$}
\STATE $w_{k+1} = w_t$
\ELSE
\STATE $w_{k+1} = w_k$
\ENDIF
\STATE Update $\delta_k$ by \autoref{alg.TR}
\IF{$|s^T(y_k - B_k s_k )| \geq \tau\| s_k\| \| y_k - B_k s_k\|$}
\STATE Update storage matrices $S_{k+1}$ and $Y_{k+1}$ by $l$ recent $\{s_j, y_j\}_{j=k-l+1}^k$
\STATE Compute $\gamma_{k+1} $ for $B_0$ by \autoref{alg.lamda2} and $\Psi_{k+1}$, $M^{-1}_{k+1}$ by \eqref{eq.dd}
\ELSE
\STATE Set $\gamma_{k+1} = \gamma_k $, $\Psi_{k+1} = \Psi_{k}$ and $M^{-1}_{k+1} = M^{-1}_k$
\ENDIF
\ENDFOR
\end{algorithmic}
\end{algorithm}
\noindent where $or = \dfrac{os}{2os + rs}$. In this work, we have considered batches corresponding to case 1. \Cref{fig1} schematically shows batches $J_k$ and $J_{k+1}$ at iterations $k$ and $k+1$, respectively, and the overlapping parts in case 1. The stochastic loss value and gradient \eqref{eq.4g} are computed at the beginning (at $w_k$) and at the end of each iteration (at trial point $w_t$). In iteration $k+1$, these quantities have to be evaluated with respect to the sample subset represented by white rectangles only. In fact, the computations with respect to subset $O_k$ at $w_{k+1}$ depend on the acceptance status of $w_t$ at iteration $k$. In case of acceptance, the loss function and gradient vector have been already computed at $w_t$; in case of rejection, these quantities are set equal to those evaluated at $w_k$ with respect to subset $O_k$. \noindent Detailed versions of \autoref{alg.lbfgs_tr_Stoc(naive)} and \autoref{alg.lsr1_tr_Stoc(naive)} are respectively provided in \autoref{alg.lbfgs_tr_Stoc} and \autoref{alg.lsr1_tr_Stoc} in \autoref{app.Algs}.


\section{Empirical study}\label{body5} 
We present in this section the results of extensive experimentation with two described stochastic QN algorithms on image classification problems. The Deep Learning Toolbox of MATLAB provides a framework for designing and implementing a deep neural network 
to perform image classification tasks using a prescribed training algorithm. Since the algorithms considered in this work, sL-BFGS-TR and sL-SR1-TR, are not defined as built-in functions, we have exploited the Deep Learning Custom Training Loops of MATLAB \footnote{\url{https://www.mathworks.com/help/deeplearning/deep-learning-custom-training-loops.html}} to implement \Cref{alg.lbfgs_tr_Stoc(naive)} and \Cref{alg.lsr1_tr_Stoc(naive)} with half-overlapping subsampling. Implementation details of the two stochastic QN algorithms considered in this work using the DL toolbox of MATLAB\footnote{\url{https://it.mathworks.com/help/deeplearning/}} are provided in 
 \url{https://github.com/MATHinDL/sL_QN_TR/} where all the codes employed to obtain the numerical results included in this paper are also available.
 
 In order to find an optimal classification model by using a $C$-class dataset, the generic problem \eqref{problem} is solved by employing the \textit{softmax cross-entropy} function $$L_i(w) = -\sum_{k=1}^{C} (y_i)_k \log (h(x_i;w))_k$$ for $i=1,\ldots, N$. One of the most popular benchmarks to make informed decisions using data-driven approaches in DL is the MNIST dataset \cite{lecun1998mnist} as $\{(x_i,y_i)\}_{i=1}^{70000}$ consisting in handwritten gray-scale images of digits $x_i$ with $28\times 28$ pixels taken values in $[0,255]$ and its corresponding labels converted to one-hot vectors. 
The Fashion-MNIST \cite{xiao2017fashion} is a variant of the original MNIST dataset which shares the same image size and structure. Its images are assigned to fashion items (clothing) belonging also to 10 classes but working with this dataset is more challenging than working with MNIST. The CIFAR-10 dataset \cite{krizhevsky2009learning} has 60000 RGB images $x_i$ of $32\times32$ pixels taken values in $[0,255]$ in 10 classes. Every single image of MNIST and Fashion-MNIST datasets is $x_i \in \mathbb{R}^{28 \times 28 \times 1}$ while one of CIFAR10 is $x_i \in \mathbb{R}^{32 \times 32 \times 3}$. In all datasets, $10000$ of the images are set aside as a testing set during training. In this work, inspired by LeNet-5 mainly used for character recognition tasks \cite{lecun1998gradient}, we have used a LeNet-like network with a shallow structure. We have also employed a modern ResNet-20 residual network \cite{he2016deep} exploiting special skip connections (shortcuts) to avoid possible gradient vanishing that might happen due to its deep architecture. Finally, we also consider a self-built convolutional neural network (CNN) named ConvNet3FC2 with a larger number of parameters than the two previous networks. In order to analyze the effect of batch normalization \cite{ioffe2015batch} on the performance of the stochastic QN algorithms, we have considered also variants of ResNet-20 and ConvNet3FC2 networks, named ResNet-20(no BN) and ConvNet3FC2(no BN), in which the batch normalization layers have been removed. \Cref{Nets} describes the networks' architecture in detail. 
 
\begin{table}
\centering
\footnotesize
\begin{tabular}{l|l}\hline\hline
\textbf{LeNet-like}  & \\\hline
Structure
& $(Conv(5\times5@20,\,1,0)\textcolor{red}{/}ReLu\textcolor{red}{/}MaxPool(2\times2,\,2,0))$\\\cline{2-2}
& $(Conv(5\times5@50,\,1,0)\textcolor{red}{/}ReLu\textcolor{red}{/}MaxPool(2\times2,\,2,0))$\\\cline{2-2}
& $FC(500\textcolor{red}{/}ReLu)$\\\cline{2-2}
& $FC(C\textcolor{red}{/}Softmax)$
\\\hline\hline
\textbf{ResNet-20}  & \\\hline
Structure     
&  $(Conv(3\times3@16,\,1,1)\textcolor{red}{/}BN\textcolor{red}{/}ReLu)$
\\\cline{2-2}
&
$
B_1
\begin{cases}
(Conv(3\times3@16,\,1,1)\textcolor{red}{/}BN\textcolor{red}{/}ReLu)\\
(Conv(3\times3@16,\,1,1)\textcolor{red}{/}BN) + addition(1)\textcolor{red}{/}Relu
\end{cases}
$
\\
&  
$
B_2
\begin{cases}
(Conv(3\times3@16,\,1,1)\textcolor{red}{/}BN\textcolor{red}{/}ReLu)\\
(Conv(3\times3@16,\,1,1)\textcolor{red}{/}BN) + addition(1)\textcolor{red}{/}Relu
\end{cases}
$
\\
& 
$
B_3
\begin{cases}
(Conv(3\times3@16,\,1,1)\textcolor{red}{/}BN\textcolor{red}{/}ReLu)\\
(Conv(3\times3@16,\,1,1)\textcolor{red}{/}BN) + addition(1)\textcolor{red}{/}Relu
\end{cases}
$
\\\cline{2-2}
& 
$
B_1
\begin{cases}
(Conv(3\times3@32,\,2,1)\textcolor{red}{/}BN\textcolor{red}{/}ReLu)\\
(Conv(3\times3@32,\,1,1)\textcolor{red}{/}BN)\\
(Conv(1\times1@32,\,2,0)\textcolor{red}{/}BN) + addition(2)\textcolor{red}{/}Relu
\end{cases}
$
\\
& 
$B_2
\begin{cases}
(Conv(3\times3@32,\,1,1)\textcolor{red}{/}BN\textcolor{red}{/}ReLu)\\
(Conv(3\times3@32,\,1,1)\textcolor{red}{/}BN) + addition(1)\textcolor{red}{/}Relu 
\end{cases}
$
\\
&
$
B_3
\begin{cases}
(Conv(3\times3@32,\,1,1)\textcolor{red}{/}BN\textcolor{red}{/}ReLu)\\
(Conv(3\times3@32,\,1,1)\textcolor{red}{/}BN) + addition(1)\textcolor{red}{/}Relu
\end{cases}
$
\\\cline{2-2}
&
$
B_1
\begin{cases}
(Conv(3\times3@64,\,2,1)\textcolor{red}{/}BN\textcolor{red}{/}ReLu)\\
(Conv(3\times3@64,\,1,1)\textcolor{red}{/}BN)\\
(Conv(1\times1@64,\,2,0)\textcolor{red}{/}BN) + addition(2)\textcolor{red}{/}Relu\\
\end{cases}
$
\\
& 
$
B_2
\begin{cases}
(Conv(3\times3@64,\,1,1)\textcolor{red}{/}BN\textcolor{red}{/}ReLu)\\
(Conv(3\times3@64,\,1,1)\textcolor{red}{/}BN) + addition(1)\textcolor{red}{/}Relu
\end{cases}
$
\\
&
$
B_3
\begin{cases}
(Conv(3\times3@64,\,1,1)\textcolor{red}{/}BN\textcolor{red}{/}ReLu)\\
(Conv(3\times3@64,\,1,1)\textcolor{red}{/}BN) + addition(1)\textcolor{red}{/}g.AvgPool\textcolor{red}{/}ReLu)
\end{cases}
$
\\\cline{2-2}
& 
$FC(C\textcolor{red}{/}Softmax)$
\\\hline\hline
\textbf{ConvNet3FC2}  &     \\\hline
Structure
& $(Conv(5\times5@32,\,1,2)\textcolor{red}{/}BN\textcolor{red}{/}ReLu\textcolor{red}{/}MaxPool(2\times2,\,1,0))$\\\cline{2-2}
& $(Conv(5\times5@32,\,1,2)\textcolor{red}{/}BN\textcolor{red}{/}ReLu\textcolor{red}{/}MaxPool(2\times2,\,1,0))$\\\cline{2-2}
& $(Conv(5\times5@64,\,1,2)\textcolor{red}{/}BN\textcolor{red}{/}ReLu\textcolor{red}{/}MaxPool(2\times2,\,1,0))$\\\cline{2-2}
& $FC(64,\textcolor{red}{/}BN\textcolor{red}{/}ReLu)$\\\cline{2-2}
& $FC(C\textcolor{red}{/}Softmax)$
\\\hline\hline
\end{tabular}
\caption{\label{Nets} Networks.}

\tabnote{\normalsize\justifying In this table,
{\small 
\begin{itemize}
    \item the syntax $(Conv(5\times5@32,\,1,2)\textcolor{red}{/}BN\textcolor{red}{/}ReLu\textcolor{red}{/}MaxPool(2\times2,\,1,0)))$ indicates a simple convolutional network (convnet) including a convolutional layer ($Conv$) using $32$ filters of size $5\times 5$, stride $1$, padding $2$,  followed by a batch normalization layer ($BN$), a nonlinear activation function ($ReLu$) and, finally, a max-pooling layer with a channel of size $2\times2$, stride 1 and padding 0.
    \item the syntax $FC(C\textcolor{red}{/}Softmax)$ indicates a layer of $C$ fully connected neurons  followed by the $softmax$ layer.
    \item the syntax $addition(1)\textcolor{red}{/}Relu$ indicates the existence of an \textit{identity shortcut} 
	    with functionality such that the output of a given block, say $B_1$ (or $B_2$ or $B_3$), is directly fed to the $addition$ layer and then to the ReLu layer while $addition(2)\textcolor{red}{/}Relu$ in a block shows the existence of a \textit{projection shortcut} by which the output from the two first convnets is added to the output of the third convnet and then the output is passed through the $ReLu$ layer.
\end{itemize}}
}
\end{table}



\begin{table}
\centering
{\footnotesize
\begin{tabular}{l|ccccc}
          & LeNet-5   & ResNet-20     & ResNet-20(no BN) & ConvNet3FC2   & ConvNet3FC2(no BN)    \\\hline\hline
MNIST     &  431,030  &  272,970      &   271,402     & 2,638,826        &      2,638,442           \\
F.MNIST   &  431,030  &  272,970      &   271,402     & 2,638,826        &      2,638,442            \\
CIFAR10   &  657,080  & 273,258       &   271,690     & 3,524,778        &      3,525,162            \\
\end{tabular}}
\caption{\label{params} The total number of networks' trainable parameters ($n$).}
\end{table}

        \begin{table}[ht]
\centering
{\footnotesize
\begin{tabular}{l|lllll}
          & LeNet-like      & ResNet-20          & ResNet-20(no BN)     & ConvNet3FC2 & ConvNet3FC2(no BN) \\\hline\hline
	MNIST     & \Centerstack[l]{\autoref{FigB} \\ Figure 8$^{\ast}$ }     &       -                                                   &       -                                                             & \Centerstack[l]{\autoref{FigB} \\ Figure 14\as}
          & \Centerstack[l]{\autoref{FigB} \\ Figure 17\as}  \\\hline
F.MNIST   & \Centerstack[l]{\autoref{FigB} \\ Figure 9\as}    & \Centerstack[l]{\autoref{FigB} \\ Figure 10\as}  
	  & \Centerstack[l]{\autoref{FigB} \\ Figure 12\as} & \Centerstack[l]{\autoref{FigB}\\ Figure 15\as}
          & \Centerstack[l]{\autoref{FigB} \\ Figure 18\as}\\\hline
CIFAR10   &     -                                           
	  & \Centerstack[l]{\autoref{FigB} \\ Figure 11\as}   & \Centerstack[l]{\autoref{FigB} \\ Figure 13\as}       
	  & \Centerstack[l]{\autoref{FigB} \\ Figure 16\as}
          & \Centerstack[l]{\autoref{FigB} \\ Figure 19\as}\\\hline
\end{tabular}}
\caption{\label{Info} Set of Figures corresponding to experiments in \Cref{Ex2}. Figures marked as $\ast$ can be found in \autoref{app.ExFig}.}
\end{table}

\subsection{Numerical Results}\label{body6.7}
\Cref{params} shows the total number of trainable parameters, $n$, for different image classification problems. We have compared sL-BFGS-TR and sL-SR1-TR in training tasks for these problems. We used hyper-parameters $c = 0.9$ and $ \tau={10}^{-2}$ in sL-BFGS-TR, $c_1 = 0.5$, $c_2 = 1.5$, $c = 10^{-6}$ and $ \tau=10^{-8}$ in sL-SR1-TR, and $\tau_1 = {10}^{-4}$, $\gamma_0 =1$, $\tau_2 =0.1$, $\tau_3 = 0.75$, $\eta_3 = 0.8$, $\eta_2=0.5$, and $\eta_4 = 2$ in both ones. We also used the same initial parameter $w_0\in \mathbb{R}^n$ by specifying the same seed to the MATLAB random number generator for both methods. All deep neural networks were trained for at most 10 epochs, and training was terminated if $100\%$ accuracy had been reached. To evaluate the performance of each model in the classification of data belonging to $C$ classes with a balanced number of samples, measuring accuracy is typically considered. The accuracy is the ratio of the number of correct predictions to the number of total predictions. In our study, we report the accuracy in percentage and overall loss values for both train and test datasets. In order to allow for better visualization, we have shown measurements of evaluation versus epochs using a determined frequency of display reported at the top of the figures. Display frequency values larger than one indicate the number of iterations that are not reported while all the iterations are considered if the display frequency is one. All figures report the results of a single run; see also additional experiments in \autoref{app.ExFig}.

We have performed extensive testing to analyze different aspects that may influence the performance of the two considered stochastic QN algorithms, mainly, the limited memory parameter and the batch size. We have also analyzed the performance of both the algorithms of interest from the point of view of CPU time. Finally, we have provided a comparison with first- and second-order methods. All experiments were performed on a Ubuntu Linux server virtual machine with 32 CPUs and 128GB RAM. 

\subsubsection{Influence of the limited memory parameter}\label{Ex1}
\begin{figure}[H]
\centering
        \includegraphics[width=0.85\linewidth, height=21cm]{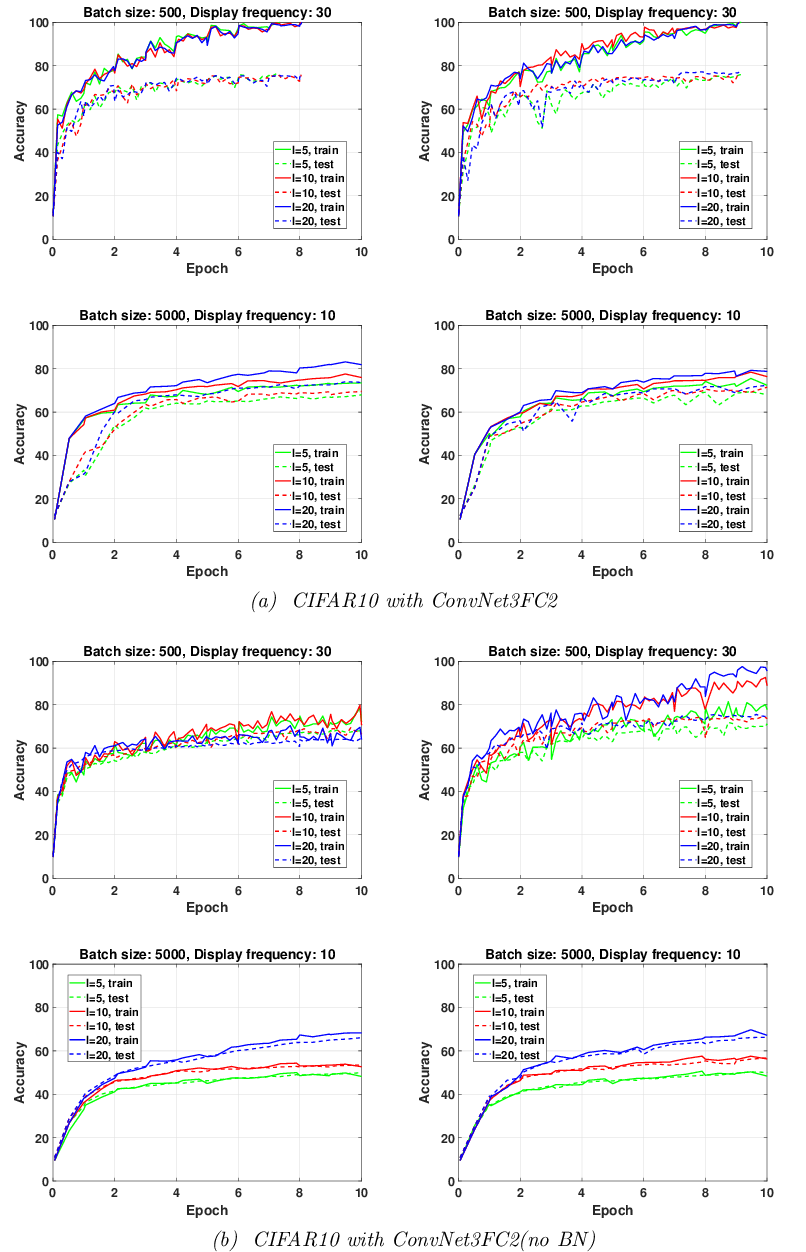}
	\caption{Performance of sL-BFGS-TR (left) and sL-SR1-TR (right) with different limited memory values ($l$).}
 \label{FigA}
\end{figure}

The results reported in \Cref{FigA} illustrate the effect of the limited memory parameter value ($l=5, 10$ and $20$) on the accuracy achieved by the two stochastic QN algorithms to train ConvNet3FC2 on CIFAR10 within a fixed number of epochs.  As it is clearly shown in this figure, in particular for ConvNet3FC2(no BN), the effect of the limited memory parameter is more pronounced when large batches are used ($bs=5000$). For large batch sizes the larger the value of $l$ the higher the accuracy. No remarkable differences in the behavior of both algorithms with small batch size ($bs=500$) are observed. It seems that incorporating more recently computed curvature vectors (i.e. larger $l$) does not increase the efficiency of the algorithms to train DNNs with BN layers while it does when BN layers are removed. Finally, we remark that we found  that using  larger values of $l$ ($l\geq 30$) was not helpful
since it led to higher over-fitting in some of our experiments.

\subsubsection{Influence of the batch size}\label{Ex2}  
\begin{figure}[H]
\centering
        \includegraphics[width=0.97\linewidth]{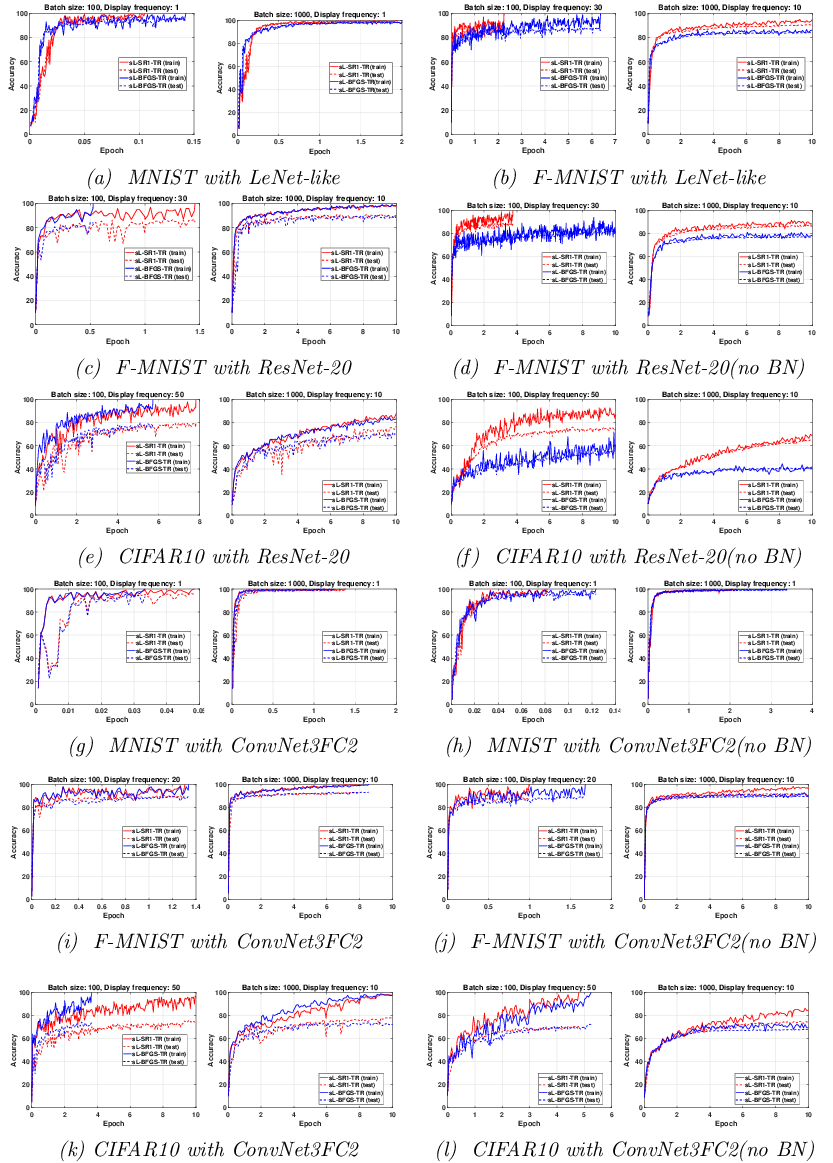}
	\caption{Evolution of the training and testing accuracy for batch sizes 100 and 1000 $(l=20)$.}
\label{FigB}
\end{figure}

We analyze the effect of the batch size on the performance of the two sQN methods while keeping fixed the limited memory parameter $l= 20$. We have considered different values of the batch size ($bs$) in $\{100,500,1000,5000\}$ or, equivalently, overlap size ($os$) in $\{50,250,500, 2500\}$ for all the problems and all the considered DNNs. The results of these experiments have been reported in \autoref{FigB} (see also Figures 8--19 in \autoref{app.ExFig}). The general conclusion is that when training the networks for a fixed number of epochs, the achieved accuracy decreases when the batch size increases. This is due to the reduction in iterations, and thus the number of parameter updates.  We have summarized in \autoref{overview} the relative superiority of one of the two algorithms over the other for all problems; "Both" refers to similar behavior. 

\autoref{overview} indicates that sL-SR1-TR  performs better than sL-BFGS-TR for training networks without BN layers while both QN updates exhibit comparable performances when used for training networks with BN layers. More detailed comments for each DNN are given below.
\begin{itemize}
\item{\textit{LeNet-like.}} The results on top of \Cref{FigB} (see also Figures 8 and 9) show that both algorithms perform well in training LeNet-like within 10 epochs to classify MNIST and Fashion-MNIST datasets, respectively. Specifically, sL-SR1-TR provides better accuracy than sL-BFGS-TR. 

\item{\textit{ResNet-20.}} \Cref{FigB} (see also Figures 10-13) shows that the classification accuracy on Fashion-MNIST increases when using ResNet-20 instead of LeNet-like, as expected. Regarding the performance of the two algorithms of interest, we see in Figures 10 and 11 that when BN is used both algorithms exhibit comparable performances. Nevertheless, we point out the fact that sL-BFGS-TR using $bs=100$ achieves higher accuracy than sL-SR1-TR in less time. Unfortunately, this comes with some awkward oscillations in the testing curves. We attribute these oscillations to a sort of inconsistency between the updated parameters and the normalized features of the testing set samples. This is due to the fact that the inference step using testing samples is done by the updated parameters and the new features which are normalized by the most recently computed moving averages of mean and variance obtained by batch normalization layers in the training phase. The numerical results on ResNet-20 without BN layers confirm this assumption can be true. These results also show that sL-SR1-TR performs better than sL-BFGS-TR in this case. Note that the experiments on  LeNet-like and ResNet-20 with and without BN layers show that sL-SR1-TR performs better than sL-BFGS-TR when batch normalization is not used, but as it can be clearly seen from the results, the elimination of BN layers causes a detriment to all method's performances. 

\item{\textit{ConvNet3FC2.}} The results of the experiments regarding this network are summarized in 
\Cref{FigB} (see also Figures 14--19). We observe also in this network some oscillations in test accuracy curves but with lower amplitude variations. The experiments show that sL-BFGS-TR still produces better testing/training accuracy than sL-SR1-TR on CIFAR10 while both algorithms behave similarly on MNIST and Fashion-MNIST datasets. Besides, sL-BFGS-TR with $bs=100$ within 10 epochs achieves the highest accuracy faster than sL-SR1-TR.
\end{itemize}
\subsubsection{CPU timings analysis}\label{Ex3}

\begin{figure}[h!]
\centering
\includegraphics[width=0.97\textwidth, height=14cm]{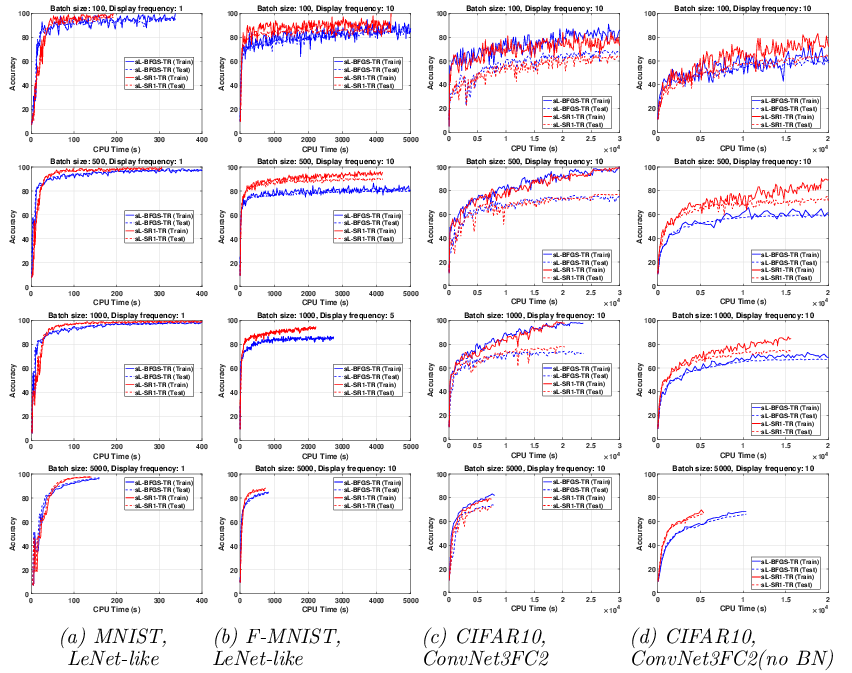}
\caption{Training accuracy vs CPU time (in seconds) of both sQN algorithms with $l=20$.}
\label{FigE}
\end{figure}

The goal is to see which algorithm achieves the highest training accuracy faster than the other one within a fixed number of epochs. \Cref{FigE} (see also Figures 20 and 21) shows that sL-SR1-TR trains faster with better accuracy than sL-BFGS-TR. We have also made a comparison of both algorithms using ConvNet3FC2 with and without BN layers. The figure shows that both algorithms 
behave comparably within the selected interval of time when BN layers are used. Nevertheless, sL-SR1-TR is faster than sL-BFGS-TR to pass 10 epochs even if it does not achieve higher training accuracy. sL-SR1-TR is also the clear winner for network models without BN layers such as ConvNet3FC2 (no BN).

\newpage

\begin{table}[H]
\centering
{
\footnotesize
\begin{tabular}{l|ccccc}
          & LeNet-5          & ResNet-20       & ResNet-20(no BN) & ConvNet3FC2   & ConvNet3FC2(no BN)    \\\hline\hline
MNIST     &   sL-SR1-TR      &   both          &   sL-SR1-TR      & both              &  both      \\ 
F.MNIST   &   sL-SR1-TR      &   both          &   sL-SR1-TR      & both              &  sL-SR1-TR \\
CIFAR10   &   sL-SR1-TR      &   sL-BFGS-TR    &   sL-SR1-TR      & sL-BFGS-TR        &  sL-SR1-TR \\
\end{tabular}}
	\caption{\label{overview} Summary of the best sQN approach for each combination problem/network architecture.}
\end{table}

This experiment illustrates that both algorithms can yield very similar training accuracy regardless of the batch size. Despite the small influence of the batch size on the final reached training and testing accuracies, it can be observed a slight increase in the accuracy when larger batch sizes are used. For this reason, one can prefer to employ larger batch sizes for sQN algorithms which can provide high benefits in view of a parallel/distributed implementation. Finally, it can be noted that based on the results of the experiments, sQN methods reveal very robust with respect to their hyper-parameters, i.e., limited memory parameter and batch size, and need minimal tuning. 

\begin{figure}[H]
\centering
\includegraphics[width=0.9\textwidth]{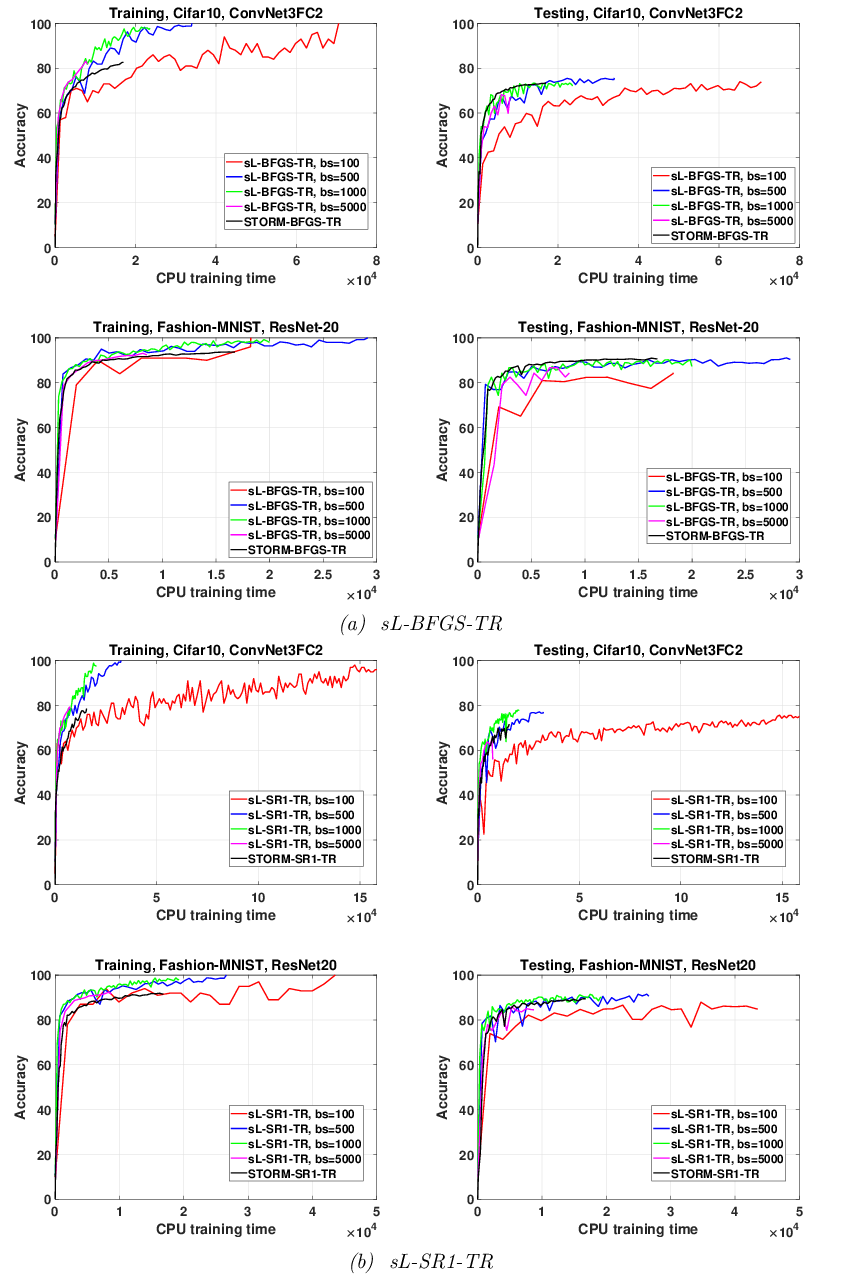}
\caption{The performance of sL-BFGS-TR and sL-SR1-TR with different fixed batch sizes ($bs$) in comparison with {\sc STORM}.}
\label{FigS}
\end{figure}

\begin{figure}[H]
\centering
\includegraphics[width=0.65\textwidth]{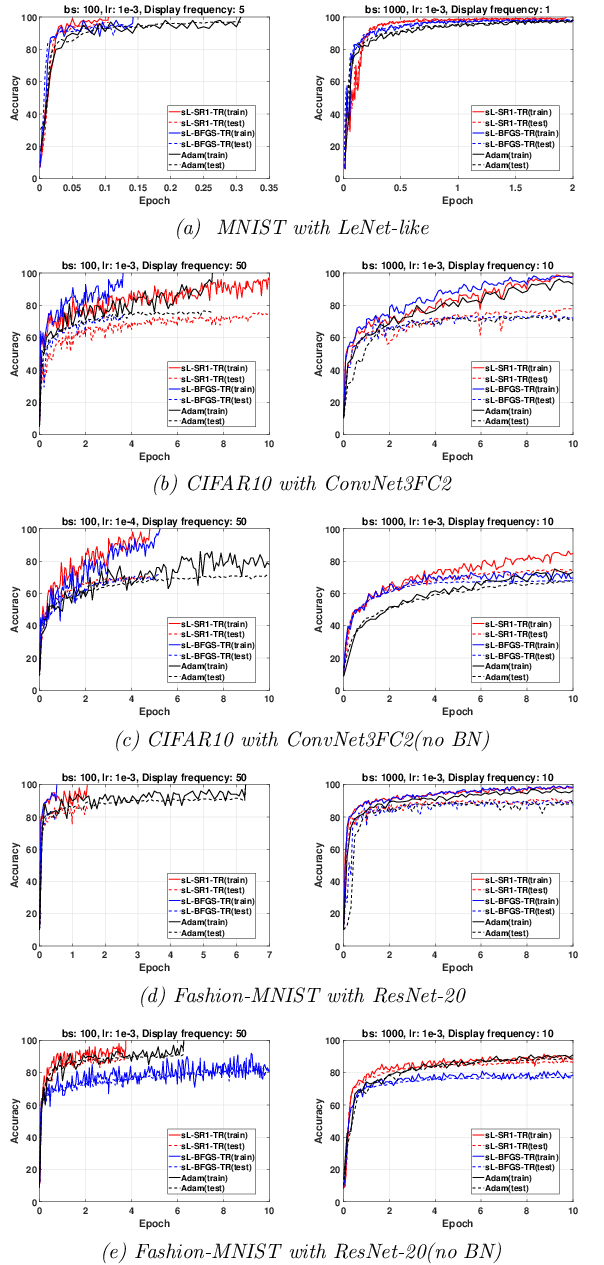}
\caption{Comparison of sL-BFGS-TR, sL-SR1-TR with $l=20$ and \textit{tuned} Adam with optimal learning rate ($lr$) for different batch sizes ($bs$).}
\label{FigH}
\end{figure}

\subsubsection{Comparison with STORM}\label{Ex3-emezzo}

We have performed a comparison of our sQN training algorithms with the algorithm {\sc STORM} (Algorithm 5 in \cite{chen2018stochastic}). {\sc STORM} relies on an adaptive batching strategy aimed at avoiding inaccurate stochastic function evaluations in the TR framework. Note that the real reduction of the objective function is not guaranteed 
in a stochastic trust-region approach. In \cite{chen2018stochastic,blanchet2019convergence}, the authors claim that if the stochastic functions are sufficiently accurate, this will increase the number of true successful iterations. Therefore, they considered a progressive sampling strategy with sample size $b_k = \min( N, \max(b_0k + b_1, \lceil\frac{1}{{\delta_k}^2}\rceil))$ where  $\delta_k$ is the trust-region radius at iteration $k$, $N$ is the total number of samples and $b_0, b_1$ are $b_0=100,\, b_1 = 32\times32\times3$ for CIFAR10 and $b_1 = 28\times28\times1$ for Fashion-MNIST. We have applied {\sc STORM} with both SR1 and BFGS updates. We have compared the performances of sL-SR1-TR and sL-BFGS-TR algorithms with different overlapping batch sizes running for $10$ epochs and {\sc STORM} with progressive batch size $b_k$ running for $50$ epochs. The largest batch size reached by {\sc STORM} was near $b_k = 25000$ (i.e., 50 percent of the total number of samples $N$). 

The results of this experiment are summarized in \Cref{FigS}. In both {Fashion-MNIST} and {CIFAR10} problems, the algorithms with $bs=500$ and $1000$ produce comparable or higher accuracy than STORM at the end of their own training phase. Even if we set a fixed budget of time corresponding to one needed for passing 50 epochs by STORM, sL-QN-TR algorithms with $bs=500$ and $1000$ provide comparable or higher accuracy. We need more consideration on the smallest and largest batch sizes. When $bs=100$, the algorithms can not be better than STORM with any fixed budgets of time; however, they provide higher training accuracy and testing accuracy, except for {Fashion-MNIST} problem on {ResNet-20} trained by sL-BFGS-TR, at the end of their training phase. This makes sense due to training with batches of small size. In contrast, when $bs=5000$, sL-BFGS-TR algorithms only can produce higher or comparable training accuracy without any comparable testing accuracy. This is normal behavior as they could update only a few parameters within 10 epochs when $bs=5000$; allowing longer training time or more epochs can compensate for this lower accuracy. This experiment also shows another finding that sL-BFGS-TR algorithms with $bs=5000$ can be preferred to $bs=100$ because they could yield higher accuracy within less time. 
\subsubsection{Comparison with Adam optimizer}\label{Ex4}
Adaptive Moment Estimation (Adam) \cite{kingma2014adam} is a popular efficient first-order optimizer used in DL.
Due to the high sensitivity of Adam to the value of its hyper-parameters, it is usually used after the determination of near-optimal values through grid searching strategies, which is a very time-consuming task. It is worth noting that sL-QN-TR approaches do not require step-length tuning, and this particular experiment offers a comparison with optimized Adam. In order to compare sL-BFGS-TR and sL-SR1-TR against Adam, we have performed a grid search of learning rates and batch sizes to select the best value of Adam's hyper-parameters. 
We consider learning rates values in $\{10^{-5},  10^{-4},  10^{-3},  10^{-2},  10^{-1}, 1\}$ and batch size in $\{100, 500, 1000, 5000\}$ and selected the values that allowed to achieve the highest testing accuracy. The gradient and squared gradient decay factors are set as $\beta_1 = 0.9$ and $\beta_2 = 0.999$, respectively. The small constant for preventing divide-by-zero errors is set to $10^{-8}$.

We have analyzed which algorithm achieves the highest training accuracy within at most 10 epochs for different batch sizes. Based on \Cref{FigH} (see also Figures 22--26), for the networks using BN layers, all methods achieve comparable training and testing accuracy within 10 epochs with $bs=1000$. However, this cannot be generally observed when $bs=100$. The figure shows \textit{tunned} Adam has higher testing accuracy than sL-SR1-TR. Nevertheless, sL-BFGS-TR is still faster to achieve the highest training accuracy, as we also previously observed, with comparable testing accuracy with \textit{tunned} Adam. On the other hand, for networks without BN layers, sL-SR1-TR is the clear winner against both other algorithms. Another important observation is that Adam is more affected by batch sizes (see Figures 22 and 23, for instance), thus the advantage over Adam can increase to enhance the parallel efficiency when using large batch sizes.

\begin{figure}[t]
\centering
\includegraphics[width=0.95\textwidth]{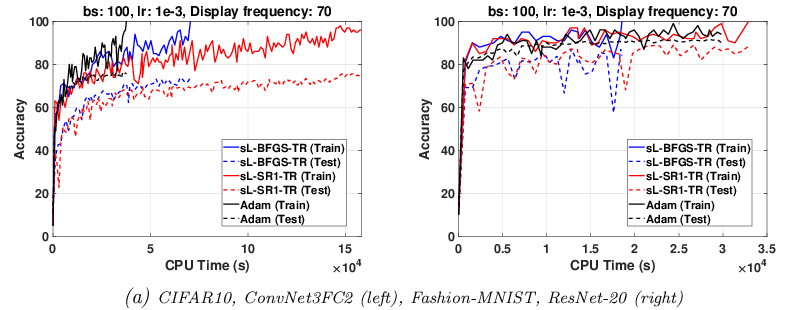}
\caption{Comparison sL-BFGS-TR, sL-SR1-TR with $l=20$ and \textit{tuned} Adam with optimal batch size ($bs$) and learning rate ($lr$) in terms of CPU training time.}
\label{time_adam}
\end{figure}

\section{Conclusions}
\label{body6}
We have studied stochastic quasi-Newton trust-region methods with L-SR1 and L-BFGS Hessian approximations for training DNNs. Extensive empirical work including the effect of batch normalization (BN), the limited memory parameter, and batch size has been reported and discussed. Our findings showed that BN is a key factor in the performance of stochastic QN algorithms and that sL-BFGS-TR behaves comparably or slightly better than sL-SR1-TR when BN layers are used while sL-SR1-TR performs better in networks without BN layers. Although this behavior is in accordance with the property of L-SR1 updates allowing for indefinite Hessian approximations in non-convex optimization, the exact reason for such different behavior is not completely clear and would deserve further investigation. Our results illustrated that employing larger batch sizes within a fixed number of epochs produces less training accuracy which can be recovered by longer training. The experiments on training time also showed a slight superiority in the accuracy reached by both algorithms when larger batch sizes are used within a fixed budget of time. This suggests the use of large batch sizes also in view of the parallelization of the algorithms. The sQN algorithms, with the overlapping fixed-size sampling strategy and fewer epochs, were more effective than the STORM algorithm which relies on a progressive adaptive sampling strategy. Finally, our results demonstrated that sQN methods are efficient in practice and, in some instances—such as when using larger batch sizes—they outperformed a tuned Adam. We believe that this contribution fills a gap concerning the real performance of the SR1 and BFGS updates onto realistic large-size DNNs and is expected to help researchers in selecting the appropriate quasi-Newton method.


\vspace{-0.2cm}
\bibliographystyle{elsarticle-num}

\newpage
\appendix
\section{Algorithms}\label{app.Algs} In this section, all algorithms considered in this work are described in detail along with the values of the hyper-parameters used in the experiments. 
\newline

\begin{minipage}{0.96\textwidth}
\begin{algorithm}[H]\caption{Overlapping multi-batch generation\label{alg:Jk}}
\begin{algorithmic}[1]\footnotesize
\begin{spacing}{1.3}
\STATE \textbf{Inputs:} $os$, $N$, $\bar{N}$, shuffled dataset and current iteration $k$
\IF{$\text{mod}(k+1, \bar{N}) \neq 0$}
\STATE Create two subsets $O_{k-1}$ and $O_k$ of size $os$
\ELSE
\IF{$\text{mod}(N,os)=0$}
\STATE Create two subsets $O_{k-1}$ and $O_k$ of size $os$ for last multi-batch $J_k$
\ELSE
\STATE Create three subsets $O_{k-1}$, $O_k$ of size $os$ and $R_k$ of size $\text{mod}(N,os)=0$ for last multi-batch $J_k$
\ENDIF
\STATE Shuffle data without replacement for the next epoch
\ENDIF
\end{spacing}
\end{algorithmic}
\end{algorithm}
\end{minipage}

\begin{minipage}{1.0\textwidth}
\begin{algorithm}[H]\caption{Trust-region radius adjustment}\label{alg.TR}
\begin{algorithmic}[1]\footnotesize
\begin{spacing}{1.3}
\STATE \textbf{Inputs:} Current iteration $k$, $\delta_k$, $\rho_k$, $0 < \tau_2 < 0.5 < \tau_3 <1$, $0 < \eta_2 \leq 0.5 $, $0.5 < \eta_3 < 1 < \eta_4$ \footnote{\,$\tau_2 =0.1$ , $\tau_3 = 0.75$, $\eta_3 = 0.8$, $\eta_2=0.5$, $\eta_4 = 2$}
\IF{$\rho_k > \tau_3$}
\IF{$\|p_k\|\leq \eta_3 \delta_k$}
\STATE $\delta_{k+1} = \delta_k$
\ELSE
\STATE $\delta_{k+1} = \eta_4\delta_k$
\ENDIF
\ELSIF{$\tau_2 \leq \rho_k \leq \tau_3$}
\STATE $\delta_{k+1} = \delta_k$
\ELSE
\STATE $\delta_{k+1} = \eta_2\delta_k$
\ENDIF
\end{spacing}
\end{algorithmic}
\end{algorithm}
\end{minipage}

\begin{minipage}{1.0\textwidth}
\begin{algorithm}[H]\caption{Orthonormal basis BFGS \label{alg:OBBFGS}}
\begin{algorithmic}[1]\footnotesize
\begin{spacing}{2.5}
\STATE \textbf{Inputs:} Current iteration $k$, $\delta \triangleq \delta_k$, $g \triangleq g_k$ and $B\triangleq B_k:\, \Psi \triangleq \Psi_k,\, M^{-1} \triangleq M^{-1}_k,\, \gamma \triangleq \gamma_k$
\STATE Compute the thin QR factors $Q$ and $R$ of $\Psi$ or the Cholesky factor $R$ of $\Psi^T\Psi$
\STATE Compute the spectral decomposition of matrix $RMR^T$, i.e., $RMR^T = U\hat{\Lambda}U^T$
\STATE Set $\hat{\Lambda} = \text{diag}(\hat{\lambda}_1, \ldots, \hat{\lambda}_k)$ such that $\hat{\lambda}_1 \leq \ldots \leq \hat{\lambda}_k$ and $\lambda_{min}=\min\{\lambda_1 , \gamma\}$ with algebraic multiplicity $r$\STATE Compute the spectral of $B_k$ as $\Lambda_1 = \hat{\Lambda}+\gamma I$
\STATE Compute $P_{\parallel}=QU$ or $P_{\parallel} = (\Psi R^{-1} U)^T$ and $g_{\parallel} = P_{\parallel}^Tg$
\IF{$\phi(0)\geq 0$}
\STATE Set: $\sigma^* = 0$
\STATE Compute $p^*$ with \eqref{eq.h} as solution of $(B_k + \sigma^* I)p = -g$
\ELSE
\STATE Compute a root $\sigma^* \in (0,\infty)$ of \eqref{eq.n} by Newton method \cite{brust2017solving}
\STATE Compute  $p^*$ with \eqref{eq.h} as solution of $(B_k + \sigma^* I)p = -g$
\ENDIF
\end{spacing}
\end{algorithmic}
\end{algorithm}
\end{minipage}

\begin{minipage}{0.9\textwidth}
\begin{algorithm}[H]\caption{L-BFGS Hessian initialization\label{alg.lamda1}}
\begin{algorithmic}[1]\footnotesize
\begin{spacing}{2}
\STATE \textbf{Inputs:} Current iteration $k$ and storage matrices $S_{k+1}$, $Y_{k+1}$, $0<1c<1$.\footnote{$c=0.9$}
\STATE Compute the smallest eigenvalue $\hat{\lambda}$ of \eqref{eq.GEV}
\IF{$\hat{\lambda}>0$}
\STATE $\gamma_{k+1} = \max\{1, c\hat{\lambda}\} \in (0,\hat{\lambda})$
\ELSE
\STATE Compute $\gamma_k^h$ by \eqref{gamma_h} and set $\gamma_{k+1} = \max \{1,\gamma_k^h\}$
\ENDIF
\end{spacing}
\end{algorithmic}
\end{algorithm}
\end{minipage}
\newpage

\begin{minipage}{0.9\textwidth}
\begin{algorithm}[H]\caption{L-BFGS-TR}\label{alg.lbfs_tr}
\begin{algorithmic}[1]\footnotesize
\begin{spacing}{1.25}
\STATE \textbf{Inputs:}
$w_0 \in \mathbb{R}^n$, $\text{epoch}_{max}$, $l$, $\gamma_0>0$, $S_0=Y_0=[\ ]$, $\delta_0>0$, $0<\tau_1, \tau<1$ \footnote{ $\text{epoch}_{max}=10$, $\gamma_0=1$, $\delta_0=1$, $\tau_1=10^{-4}$, $\tau=10^{-2}$}
\STATE Compute $f_0$ and $g_0$ by \eqref{detCompu}
\FOR{$k = 0, 1,\dots$}
\IF{$\text{mod}(k+1,\bar{N}) = 0$}
\STATE {Shuffle the data without replacement for the next epoch and $\text{epoch} = \text{epoch} + 1$}
\ENDIF
\STATE \hfill\COMMENT{Check exit condition}
\IF{$\text{epoch} > \text{epoch}_{max}$}
\STATE \textbf{Stop training}
\ENDIF
\STATE \hfill\COMMENT{Compute $p\triangleq p^{*}_k$}
\IF{$k=0$} 
\STATE Compute $p = - \delta_k\dfrac{g_k}{\|g_k\|}$
\ELSE 
\STATE Compute $p$ using \autoref{alg:OBBFGS}
\ENDIF
\STATE \hfill\COMMENT{Compute trial $w_t$}
\STATE Compute $w_{t} = w_k + p$ and then $f_{t}$ and $g_{t}$ by \eqref{detCompu}
\STATE \hfill\COMMENT{Curvature pair and $\rho_k$}
\STATE Compute $(s_k , y_k) = (w_{t}-w_k, g_{t}-g_k)$ and $\rho_k = \dfrac{f_t - f_k}{Q(p)}$
\STATE \hfill\COMMENT{Update $w_k$}
\IF{$\rho_k \geq \tau_1$}
\STATE $w_{k+1}=w_t$
\ELSE
\STATE $w_{k+1}=w_k$
\ENDIF
\STATE \hfill\COMMENT{Update $\delta_k$}
\STATE Update $\delta_k$ by \autoref{alg.TR}
\STATE \hfill\COMMENT{Update $B_k$} 
\IF{$s_k^Ty_k>\tau\|s_k\|^2$}
\IF{$k < l$}
\STATE Store $s_k$ and $y_k$ as new columns in $S_{k+1}$ and $Y_{k+1}$
\ELSE
\STATE Keep only $l$ most recent $\{s_j, y_j\}_{j=k-l+1}^k$ in $S_{k+1}$ and $Y_{k+1}$
\ENDIF
\STATE Compute $\gamma_{k+1} $ for $B_0$ by \autoref{alg.lamda1} and $\Psi_{k+1}$, $M^{-1}_{k+1}$ by \eqref{eq.b}
\ELSE
\STATE Set $\gamma_{k+1} = \gamma_k $, $\Psi_{k+1} = \Psi_{k}$ and $M^{-1}_{k+1} = M^{-1}_k$
\ENDIF
\ENDFOR
\end{spacing}
\end{algorithmic}
\end{algorithm}
\end{minipage}

\begin{minipage}{0.96\textwidth}
\begin{algorithm}[H]\caption{Orthonormal Basis SR1 (OBS)\label{alg:OBS}}
\begin{algorithmic}[1]\footnotesize
\begin{spacing}{1.7}
\STATE \textbf{Inputs:} Current iteration $k$, $\delta \triangleq \delta_k$, $g \triangleq g_k$ and $B\triangleq B_k:\, \Psi \triangleq \Psi_k,\, M^{-1} \triangleq M^{-1}_k,\,\gamma \triangleq \gamma_k$
\STATE Compute the thin QR factors $Q$ and $R$ of $\Psi$ or the Cholesky factor $R$ of $\Psi^T\Psi$
\STATE Compute the spectral decomposition of matrix $RMR^T$, i.e., $RMR^T = U\hat{\Lambda}U^T$
\STATE Set $\hat{\Lambda} = \text{diag}(\hat{\lambda}_1, \ldots, \hat{\lambda}_k)$ s.t. $\hat{\lambda}_1 \leq \ldots \leq \hat{\lambda}_k$ and $\lambda_{min}=\min\{\lambda_1 , \gamma\}$ with algebraic multiplicity $r$
\STATE Compute the spectral of $B_k$ as $\Lambda_1 = \hat{\Lambda}+\gamma I$
\STATE Compute $P_{\parallel}=QU$ or $P_{\parallel} = (\Psi R^{-1} U)^T$ and $g_{\parallel} = P_{\parallel}^Tg$
\STATE \hfill \COMMENT{feasible constraint: $\phi(-\lambda_{min})\geq 0$}
\IF{Case I: $\lambda_{min}>0$ and $\phi(0)\geq 0$}
\STATE Set: $\sigma^* = 0$
\STATE Compute $p^*$ with \eqref{eq.h} as solution of $(B_k + \sigma^* I)p =-g$
\ELSIF{Case II: $\lambda_{min} \leq 0$ and $\phi(-\lambda_{min}) \geq 0$}
\STATE Set: $\sigma^* = -\lambda_{min}$
\STATE Compute $p^*$ with \eqref{eq.gg} as solution of $(B_k + \sigma^* I)p= -g$
\IF{Case III: $\lambda_{min}<0$}
\STATE Compute $\alpha$ and $u_{min}$ with \eqref{eq.ii} for $z^* = \alpha u_{min}$
\STATE Update: $p^* = p^* + z^*$
\ENDIF
\STATE \hfill \COMMENT{infeasible constraint: $\phi(-\lambda_{min}) < 0$}
\ELSE
\STATE Compute a root $\sigma^* \in (\max\{-\lambda_{min}, 0\},\infty)$ of \eqref{eq.n} by Newton method \cite{brust2017solving}
\STATE Compute  $p^*$ with \eqref{eq.h} as solution of $(B_k + \sigma^* I)p = -g$
\ENDIF
\end{spacing}
\end{algorithmic}
\end{algorithm}
\end{minipage}

\begin{minipage}{0.9\textwidth}
\begin{algorithm}[H]\caption{L-SR1 Hessian initialization\label{alg.lamda2}}
\begin{algorithmic}[1]\footnotesize
\begin{spacing}{1.4}
\STATE \textbf{Inputs:} Current iteration $k$ and storage matrices $S_{k+1}$, $Y_{k+1}$  \footnote{$c_1 = 0.5$, $c_2 = 1.5$, $c = 10^{-6}$}
\STATE Compute the smallest eigenvalue $\hat{\lambda}$ of \eqref{eq.GEV}
\IF{$\hat{\lambda}>0$}
\STATE $\gamma_{k+1} = \max\{c, c_1\hat{\lambda}\}$
\ELSE
\STATE $\gamma_{k+1} = \min\{ -c, c_2 \hat{\lambda}\}$
\ENDIF
\end{spacing}
\end{algorithmic}
\end{algorithm}
\end{minipage}
\newpage

\newpage
\vspace{-2mm}
\begin{minipage}{0.9\textwidth}
\begin{algorithm}[H]\caption{L-SR1-TR}\label{alg.lsr1_tr}
\begin{algorithmic}[1]\footnotesize
\begin{spacing}{1.25}
\STATE \textbf{Inputs:}
$w_0 \in \mathbb{R}^n$, $\text{epoch}_{max}$, $l$, $\gamma_0>0$, $S_0=Y_0=[\ ]$, $\delta_0>0$, $0<\tau_1, \tau<1$ \footnote{ $\text{epoch}_{max}=10$, $\gamma_0=1$, $\delta_0=1$, $\tau_1=10^{-4}$, $\tau=10^{-8}$}
\STATE Compute $f_0$ and $g_0$ by \eqref{detCompu}
\FOR{$k = 0, 1,\dots$}
\IF{$\text{mod}(k+1,\bar{N}) = 0$}
\STATE {Shuffle the data without replacement for the next epoch and $\text{epoch} = \text{epoch} + 1$}
\ENDIF
\STATE \hfill\COMMENT{Check exit condition}
\IF{$\text{epoch} > \text{epoch}_{max}$}
\STATE \textbf{Stop training}
\ENDIF
\STATE \hfill\COMMENT{Compute $p\triangleq p^{*}_k$}
\IF{$k=0$} 
\STATE Compute $p = - \delta_k\dfrac{g_k}{\|g_k\|}$
\ELSE 
\STATE Compute $p$ using \autoref{alg:OBS}
\ENDIF
\STATE \hfill\COMMENT{Compute trial $w_t$}
%
\STATE Compute $w_{t} = w_k + p$ and then $f_{t}$ and $g_{t}$ by \eqref{detCompu}
\STATE \hfill\COMMENT{Curvature pair and $\rho_k$}
%
\STATE Compute $(s_k , y_k) = (w_{t}-w_k, g_{t}-g_k)$ and $\rho_k = \dfrac{f_t - f_k}{Q(p)}$
\STATE \hfill\COMMENT{Update $w_k$}
%
\IF{$\rho_k \geq \tau_1$}
\STATE $w_{k+1}=w_t$
\ELSE
\STATE $w_{k+1}=w_k$
\ENDIF
\STATE \hfill\COMMENT{Update $\delta_k$}
\STATE Update: $\delta_k$ with \autoref{alg.TR}
\STATE \hfill\COMMENT{Update $B_k$} 
\IF{$|s^T(y_k - B_k s_k )| \geq \tau\| s_k\| \| y_k - B_k s_k\|$}
\IF{$k < l$}
\STATE Store $s_k$ and $y_k$ as new columns in $S_{k+1}$ and $Y_{k+1}$
\ELSE
\STATE Keep only $l$ most recent $\{s_j, y_j\}_{j=k-l+1}^k$ in $S_{k+1}$ and $Y_{k+1}$
\ENDIF
\STATE Compute $\gamma_{k+1} $ for $B_0$ by \autoref{alg.lamda2} and $\Psi_{k+1}$, $M^{-1}_{k+1}$ by \eqref{eq.dd}
\ELSE
\STATE Set $\gamma_{k+1} = \gamma_k $, $\Psi_{k+1} = \Psi_{k}$ and $M^{-1}_{k+1} = M^{-1}_k$
\ENDIF
\ENDFOR
\end{spacing}
\end{algorithmic}
\end{algorithm}
\end{minipage}

\begin{minipage}{0.94\textwidth}
\begin{algorithm}[H]\caption{sL-BFGS-TR}\label{alg.lbfgs_tr_Stoc}
\begin{algorithmic}[1]\small
\STATE \textbf{Inputs:} $w_0 \in \mathbb{R}^n$, $\text{epoch}_{max}$, $l$, $\gamma_0>0$, $S_0=Y_0=[\ ]$, $\delta_0>0$, $0<\tau_1, \tau<1$ \footnote{ $\text{epoch}_{max}=10$, $\gamma_0=1$, $\delta_0=1$, $\tau_1=10^{-4}$, $\tau=10^{-2}$}
\WHILE{True}
\IF{$k = 0$}
\STATE Take first and second subsets $O_{-1}$ and $O_{0}$ of size $os$ for the initial multi-batch $J_0$
\STATE Compute $f_0^{O_{-1}}$, $g_0^{O_{-1}}$ and $f_0^{O_{0}}$, $g_0^{O_{0}}$ by \eqref{eq.4a} and then $f_0^{J_0}$, $ g_0^{J_0}$ by \eqref{eq.4g}
\ELSE
\STATE Take the second subset $O_k$ of size $os$ for the multi-batch $J_k$
\STATE Compute $f_k^{O_k}$, $g_k^{O_k}$ by \eqref{eq.4a}, and then $f_k^{J_k}$, $ g_k^{J_k}$ by \eqref{eq.4g}
\IF{$\text{mod}(k+1,\bar{N}) =  0$}
\STATE {Shuffle the data without replacement for the next epoch and $\text{epoch} = \text{epoch} + 1$}
\ENDIF
\ENDIF
\STATE \hfill\COMMENT{Check exit condition}
\IF{$\text{epoch} > \text{epoch}_{max}$}
\STATE \textbf{Stop training}
\ENDIF
\STATE \hfill\COMMENT{Compute search direction}
\IF{$k=0$}
\STATE Compute $p_k = - \delta_k\dfrac{g_k^{J_k}}{\|g_k^{J_k}\|}$
\ELSE
\STATE Compute $p_k$ using \autoref{alg:OBBFGS}
\ENDIF
\STATE \hfill\COMMENT{Compute trial $w_t$}
\STATE Compute $w_t = w_k + p_k$
\STATE Compute $f_{t}^{O_{k-1}}$, $g_{t}^{O_{k-1}}$ and $f_{t}^{O_{k}}$, $g_{t}^{O_k}$ by \eqref{eq.4a} and then $f_{t}^{J_k}$, $g_{t}^{J_k}$ by \eqref{eq.4g}
\STATE \hfill\COMMENT{Compute curvature pair and $\rho_k$}
\STATE Compute $(s_k, y_k) = (w_{t}-w_{k}, g_{t}^{J_k} - g_{k}^{J_k})$ and $\rho_k = \dfrac{f^{J_k}_{t} - f^{J_k}_k}{Q(p_k)}$ 
\IF{$\rho_k \geq \tau_1$} 
\STATE $w_{k+1} = w_t$\hfill\COMMENT{Update $w_k$}
\ELSE 
\STATE $w_{k+1} = w_k$
\ENDIF
\STATE Update $\delta_k$ by \autoref{alg.TR}\hfill\COMMENT{Update $\delta_k$}
\IF{$s_k^Ty_k > \tau \|s_k\|^2$}
\IF{$k < l$}
\STATE Store $s_k$ and $y_k$ as new columns in $S_{k+1}$ and $Y_{k+1}$\hfill\COMMENT{Update $B_k$}
\ELSE
\STATE Keep only $l$ recent $\{s_j, y_j\}_{j=k-l+1}^k$ in $S_{k+1}$ and $Y_{k+1}$
\ENDIF
\STATE Compute $\gamma_{k+1} $ for $B_0$ by \autoref{alg.lamda1} and $\Psi_{k+1}$, $M^{-1}_{k+1}$ by \eqref{eq.b}
\ELSE
\STATE Set $\gamma_{k+1} = \gamma_k $, $\Psi_{k+1} = \Psi_{k}$ and $M^{-1}_{k+1} = M^{-1}_k$
\ENDIF
\STATE $k = k+1$
\ENDWHILE
\end{algorithmic}
\end{algorithm}
\end{minipage}

\begin{minipage}{0.94\textwidth}
\begin{algorithm}[H]\caption{sL-SR1-TR}\label{alg.lsr1_tr_Stoc}
\begin{algorithmic}[1]\small
\STATE \textbf{Inputs:} $w_0 \in \mathbb{R}^n$, $\text{epoch}_{max}$, $l$, $\gamma_0>0$, $S_0=Y_0=[\ ]$, $\delta_0>0$, $0<\tau_1, \tau<1$ \footnote{ $\text{epoch}_{max}=10$, $\gamma_0=1$, $\delta_0=1$, $\tau_1=10^{-4}$, $\tau=10^{-8}$}
\WHILE{True}
\IF{$k = 0$}
\STATE Take first and second subsets $O_{-1}$ and $O_{0}$ of size $os$ for the initial multi-batch $J_0$
\STATE Compute $f_0^{O_{-1}}$, $g_0^{O_{-1}}$ and $f_0^{O_{0}}$, $g_0^{O_{0}}$ by \eqref{eq.4a} and then $f_0^{J_0}$, $ g_0^{J_0}$ by \eqref{eq.4g}
\ELSE
\STATE Take the second subset $O_k$ of size $os$ for the multi-batch $J_k$
\STATE Compute $f_k^{O_k}$, $g_k^{O_k}$ by \eqref{eq.4a}, and then $f_k^{J_k}$, $ g_k^{J_k}$ by \eqref{eq.4g}
\IF{$\text{mod}(k+1,\bar{N}) =  0$}
\STATE {Shuffle the data without replacement for the next epoch and $\text{epoch} = \text{epoch} + 1$}
\ENDIF
\ENDIF
\STATE \hfill\COMMENT{Check exit condition}
\IF{$\text{epoch} > \text{epoch}_{max}$}
\STATE \textbf{Stop training}
\ENDIF
\STATE \hfill\COMMENT{Compute search direction}
\IF{$k=0$}
\STATE Compute $p_k = - \delta_k\dfrac{g_k^{J_k}}{\|g_k^{J_k}\|}$
\ELSE
\STATE Compute $p_k$ using \autoref{alg:OBBFGS}
\ENDIF
\STATE \hfill\COMMENT{Compute trial $w_t$}
\STATE Compute $w_t = w_k + p_k$
\STATE Compute $f_{t}^{O_{k-1}}$, $g_{t}^{O_{k-1}}$ and $f_{t}^{O_{k}}$, $g_{t}^{O_k}$ by \eqref{eq.4a} and then $f_{t}^{J_k}$, $g_{t}^{J_k}$ by \eqref{eq.4g}
\STATE \hfill\COMMENT{Compute curvature pair and $\rho_k$}
\STATE Compute $(s_k, y_k) = (w_{t}-w_{k}, g_{t}^{J_k} - g_{k}^{J_k})$ and $\rho_k = \dfrac{f^{J_k}_{t} - f^{J_k}_k}{Q(p_k)}$ 
\IF{$\rho_k \geq \tau_1$}
\STATE $w_{k+1} = w_t$\hfill\COMMENT{Update $w_k$}
\ELSE
\STATE $w_{k+1} = w_k$
\ENDIF
\STATE Update $\delta_k$ by \autoref{alg.TR} \hfill\COMMENT{Update $\delta_k$}
\IF{$|s^T(y_k - B_k s_k )| \geq \tau\| s_k\| \| y_k - B_k s_k\|$}
\IF{$k \leq l$}
\STATE Store $s_k$ and $y_k$ as new columns in $S_{k+1}$ and $Y_{k+1}$\hfill\COMMENT{Update $B_k$}
\ELSE
\STATE Keep only $l$ recent $\{s_j, y_j\}_{j=k-l+1}^k$ in $S_{k+1}$ and $Y_{k+1}$
\ENDIF
\STATE Compute $\gamma_{k+1} $ for $B_0$ by \autoref{alg.lamda2} and $\Psi_{k+1}$, $M^{-1}_{k+1}$ by \eqref{eq.dd}
\ELSE
\STATE Set $\gamma_{k+1} = \gamma_k $, $\Psi_{k+1} = \Psi_{k}$ and $M^{-1}_{k+1} = M^{-1}_k$
\ENDIF
\STATE $k = k+1$
\ENDWHILE
\end{algorithmic}
\end{algorithm}
\end{minipage}

\section{Solving the Trust-Region subproblem}\label{app.TRsolv} 
\subsection{Computing the search direction in the L-BFGS-TR method}

We describe in this subsection how to solve the trust-region subproblem \eqref{eq.TRsub} where the BFGS Hessian approximation $B_k$ is in compact form; see  \cite{adhikari2017limited, brust2017solving, rafati2018improving} for more details.

Let $B_k$ be an L-BFGS compact matrix \eqref{eq.b}. Using \autoref{Theorem1}, the global solution of the trust-region subproblem \eqref{eq.TRsub} can be obtained by exploiting the following two strategies:

\noindent
\paragraph*{Spectral decomposition of $B_k$} Computing the thin QR factorization of matrix $\Psi_k$,  $\Psi_k = Q_kR_k$, or the Cholesky factorization of $\Psi_k^T\Psi_k$,  $\Psi_k^T\Psi_k = R^TR$, and then spectrally decomposing the small matrix $R_kM_kR_k^T$ as $R_kM_kR_k^T = U_k\hat{\Lambda}U_k^T$ leads to
$$B_k = B_0 + Q_kR_kM_kR_k^TQ_k^T = \gamma_k I + Q_kU_k\hat{\Lambda}U_k^TQ_k^T,$$
where $U_k$ and $\hat{\Lambda}$ are orthogonal and diagonal matrices, respectively. Let $P_{\parallel} \triangleq Q_kU_k$ (or $P_{\parallel} = (\Psi_kR_k^{-1}U_k)^T$) and $P_{\perp} \triangleq (Q_kU_k)^\perp$ where $(.)^{\perp}$ denotes orthogonal complement. By Theorem 2.1.1 in \cite{golub2013matrix}, we have $P^TP = PP^T = I$ where
\begin{equation}\label{eq.e}
    P \triangleq \begin{bmatrix} P_{\parallel}& P_{\perp}\end{bmatrix}\in \mathbb{R}^{n\times n}.
\end{equation}
Therefore the spectral decomposition of $B_k$ is obtained as 
\begin{equation}\label{eq.f}
    B_k = P\Lambda P^T,\qquad
    \Lambda \triangleq
    \begin{bmatrix}
    \Lambda_1 & 0\\
    0 & \Lambda_2
    \end{bmatrix}
    =
    \begin{bmatrix}
    \hat{\Lambda} + \gamma_k I & 0\\
    0 & \gamma_k I
    \end{bmatrix},
\end{equation}
where $\Lambda_1$ consists of at most $2l$ eigenvalues as $\Lambda_1  = \diag(\hat{\lambda}_1 +\gamma_k , \hat{\lambda}_2+\gamma_k, \dots, \hat{\lambda}_{2l}+\gamma_k)$. We assume the eigenvalues are increasingly ordered.

\noindent
\paragraph*{Inversion by Sherman-Morrison-Woodbury formula} By dropping subscript $k$ in \eqref{eq.b} and using the Sherman-Morrison-Woodbury formula to compute the inverse of the coefficient matrix in \eqref{eq.opt}, we have
\begin{equation}\label{eq.h}
    p(\sigma) = -(B + \sigma I)^{-1}g = -\frac{1}{\tau} \left( I - \Psi\left(\tau M^{-1}+\Psi^T\Psi\right)^{-1}\Psi^T \right)g,
\end{equation}
where $\tau = \gamma + \sigma$. By using \eqref{eq.f}, the first optimality condition in \eqref{eq.opt} can be written as
\begin{equation}\label{eq.i}
    (\Lambda + \sigma I)v = -P^Tg,
\end{equation}
where 
\begin{equation}\label{eq.j}
     v  = P^Tp,\qquad
     P^Tg  \triangleq
     \begin{bmatrix}
     g_{\parallel}\\g_{\perp}
     \end{bmatrix} = 
     \begin{bmatrix}
     P_{\parallel}^Tg\\P_{\perp}^Tg
     \end{bmatrix},
\end{equation}
and therefore
\begin{equation}\label{eq.k}
\| p(\sigma)\| = \| v(\sigma)\| =
\sqrt{ \left\{ \sum_{i=1}^{k} {\dfrac{(g_{\parallel})_i^2}{(\lambda_i + \sigma)^2}} \right\} + \dfrac{\|g_{\perp}\|^2}{(\gamma + \sigma)^2}},
\end{equation}
where $\|g_{\perp}\|^2 = \|g\|^2 - \|g_{\parallel}\|^2$. This makes the computation of $\|p\|$ feasible without computing $p$ explicitly. Let $p_u \triangleq p(0)$ as an unconstrained minimizer for \eqref{eq.TRsub} be the solution of the first optimality condition in \eqref{eq.opt}, for which $\sigma = 0$ makes the second optimality condition hold. Now, we consider the following cases:
\begin{itemize}
    \item If $\| p_u\| \leq \delta$, the optimal solution of \eqref{eq.TRsub} using \eqref{eq.h} is computed as
    \begin{equation}\label{eq.m}
        (\sigma^*, p^*) = (0, p_u) = (0, p(0)).
    \end{equation}
    \item If $\| p_u\| >\delta$, then $p^*$ must lie on the boundary of the trust-region to hold the second optimality condition. To impose this, $\sigma^*$ must be the root of the following equation which is determined by the Newton method proposed in \cite{brust2017solving}:
    \begin{equation}\label{eq.n}
    \phi(\sigma) \triangleq \dfrac{1}{\|p(\sigma)\|}-\dfrac{1}{\delta} = 0.
    \end{equation}
    Therefore, using \eqref{eq.h}, the global solution is computed as
    \begin{equation}\label{eq.o}
        (\sigma^*, p^*) =  (\sigma^*, p(\sigma^*)).
    \end{equation}
\end{itemize}
The procedure described in this section to solve the trust-region subproblem is illustrated in \autoref{alg:OBBFGS} (see \autoref{app.Algs}). 

\subsection{Computing the search direction in the L-SR1-TR method}
To solve \eqref{eq.TRsub} 
 where $B_k$ is a compact L-SR1 matrix \eqref{eq.bb}, an efficient algorithm called the \textit{Orthonormal Basis L-SR1} (OBS) was proposed in \cite{brust2017solving}. We summarize this approach here.

Let \eqref{eq.f} be the eigenvalue decomposition of \eqref{eq.bb}, where $\Lambda_1$ consists of at most $l$ eigenvalues as $\Lambda_1  = \diag(\hat{\lambda}_1 +\gamma_k , \hat{\lambda}_2+\gamma_k, \dots, \hat{\lambda}_{l}+\gamma_k)$. We assume the eigenvalues are increasingly ordered. The OBS method exploits the Sherman-Morrison-Woodbury formula in different cases for L-SR1 $B_k$; by dropping subscript $k$ in \eqref{eq.bb}, these cases are: 

\noindent
\paragraph*{$B$ is positive definite} In this case, the global solution of \eqref{eq.TRsub} is \eqref{eq.m} or \eqref{eq.o}.
\noindent
\paragraph*{$B$ is positive semi-definite (singular)} Since $\gamma \neq 0$ and $B$ is positive semi-definite with all non-negative eigenvalues, then $\lambda_{min} = \min\{ \lambda_1, \gamma\} = \lambda_1 = 0$. Let $r$ be the multiplicity of the $\lambda_{min}$; therefore,
$$0=\lambda_1 = \lambda_2 =\dots = \lambda_r<\lambda_{r+1}\leq \lambda_{r+2}\leq \dots \leq \lambda_{k}.$$
For $\sigma > -\lambda_{min}=0$, the matrix $(\Lambda + \sigma I)$ in \eqref{eq.i} is invertible, and thus, $\|p(\sigma)\|$ in \eqref{eq.k} is well-defined. For $\sigma=-\lambda_{min}=0$, we consider the two following sub-cases\footnote{To have a well-defined expression in \eqref{eq.k}, we will discuss in limit setting (at $-\lambda_{min}^{+}$).}: 
\begin{enumerate}
	\item If $\lim_{\sigma \to 0^{+}} \phi(\sigma) <0$, then $\lim_{\sigma \to 0^{+}} \| p(\sigma)\| > \delta$. Here, the OBS algorithm uses Newton's method to find $\sigma^* \in (0, \infty)$ so that the global solution $p^*$ lies on the boundary of trust-region, i.e., $\phi(\sigma^*) = 0$. This
    solution $p^*=p(\sigma^*)$ is computed using \eqref{eq.h}; by that, the global pair solution $(\sigma^*, p^*)$ satisfies the first and second optimal conditions in \eqref{eq.opt}.
    \item If $\lim_{\sigma \to 0^{+}} \phi(\sigma) \geq 0$, then $\lim_{\sigma \to 0^{+}} \| p(\sigma)\| \leq \delta$. It can be proved that $\phi(\sigma)$ is strictly increasing for $\sigma > 0$ (see Lemma 7.3.1 in \cite{conn2000trust}). This makes $\phi(\sigma) \geq 0$ for $\sigma > 0$ as it is non-negative at $0^{+}$, and thus, $\phi(\sigma)$ can only have a root $\sigma^* = 0$ in $\sigma \geq 0$. Here, we should notice that even if $\phi(\sigma)>0$, the solution $\sigma^*=0$ makes the second optimality condition in \eqref{eq.opt} hold. Since matrix $B + \sigma I$ at $\sigma^*=0$ is not invertible, the global solution $p^*$ for the first optimality condition in \eqref{eq.opt} is computed by
    \begin{equation}\label{eq.gg}
    \begin{split}
        p^* & = p(\sigma^*) = -(B + \sigma^* I)^{\dagger}g = -P(\Lambda + \sigma^* I)^{\dagger}P^T g\\
        & = -P_{\parallel}(\Lambda_1 + \sigma^* I)^{\dagger}P_{\parallel}^Tg - \dfrac{1}{\gamma + \sigma^*}P_{\perp}P_{\perp}^Tg\\
        & = -\Psi R^{-1}U(\Lambda_1 + \sigma^* I)^{\dagger}g_{\parallel} - \dfrac{1}{\gamma + \sigma^*}P_{\perp}P_{\perp}^Tg,
    \end{split}
    \end{equation}
    where $(g_{\parallel})_i = (P_{\parallel}^Tg)_i = 0$ for $i=1,\dots, r$ if $\sigma^* = -\lambda_{min}=-\lambda_1=0$, and
    $$P_{\perp}P_{\perp}^Tg = (I - P_{\parallel}P_{\parallel}^T)g = (I-\Psi R^{-1}R^{-T}\Psi^T)g.$$
    Therefore, both optimality conditions in \eqref{eq.opt} hold for the pair solution $(\sigma^*, p^*)$.
\end{enumerate}
\paragraph*{$\mathbf{B}$ is indefinite} Let $r$ be the algebraic multiplicity of the leftmost eigenvalue $\lambda_{min}$. Since $B$ is indefinite and $\gamma \neq 0$, we have $\lambda_{min} = \min\{ \lambda_1, \gamma\} < 0.$\\\noindent
Obviously, for $\sigma > -\lambda_{min}$, the matrix $(\Lambda + \sigma I)$ in \eqref{eq.i} is invertible, and thus, $\|p(\sigma)\|$ in \eqref{eq.k} is well-defined. For $\sigma=-\lambda_{min}$, we discuss the two following cases:
\begin{enumerate}
    \item If $\lim_{\sigma \to -\lambda_{min}^{+}} \phi(\sigma) <0$,  then $\lim_{\sigma \to -\lambda_{min}^{+}} \| p(\sigma)\| > \delta$. The OBS algorithm uses Newton's method to find $\sigma^* \in (-\lambda_{min},\infty)$ as the root of $\phi(\sigma)=0$ so that the global solution $p^*$ lies on the boundary of trust-region. By using \eqref{eq.h} to compute $p^*=p(\sigma^*)$, the pair $(\sigma^*, p^*)$ satisfies the both conditions in \eqref{eq.opt}. 
    \item If $\lim_{\sigma \to -\lambda_{min}^{+}} \phi(\sigma) \geq 0$, then $\lim_{\sigma \to -\lambda_{min}^{+}} \| p(\sigma)\| \geq \delta$. For $\sigma > -\lambda_{min}$, we have $\phi(\sigma) \geq 0$ but the solution $\sigma^* = -\lambda_{min}$ as the only root of $\phi(\sigma) = 0$ is a positive number, which cannot satisfy the second optimal condition when $\phi(\sigma)$ is strictly positive. Hence, we should consider the cases of equality and inequality separately:
     \\
		\textbf{Equality.} Let $\lim_{\sigma \to -\lambda_{min}^{+}} \phi(\sigma) = 0$. Since matrix $B + \sigma I$ at $\sigma^*=-\lambda_{min}$ is not invertible, the global solution $p^*$ for the first optimality condition in \eqref{eq.opt} is computed using \eqref{eq.gg} by
        \begin{equation}\label{eq.hh}
            p^* 
		 = 
            \begin{cases}
            -\Psi R^{-1}U(\Lambda_1 + \sigma^* I)^{\dagger}g_{\parallel} - \dfrac{1}{\gamma + \sigma^*}P_{\perp}P_{\perp}^Tg,
            &
            \qquad \sigma^* \neq  -\gamma,\\
            -\Psi R^{-1}U(\Lambda_1 + \sigma^* I)^{\dagger}g_{\parallel},
            &
            \qquad \sigma^* = -\gamma,
            \end{cases}
        \end{equation}
    where $g_{\perp} = P_{\perp}^Tg = 0$, and thus $\|g_{\perp}\| = 0$ if $\sigma^* = -\lambda_{min}=-\gamma$. Moreover, $(g_{\parallel})_i = (P_{\parallel}^Tg)_i = 0$ for $i=1,\dots, r$ if $\sigma^* = -\lambda_{min}=-\lambda_1$. 

    \noindent We note that both optimality conditions in \eqref{eq.opt} hold for the computed $(\sigma^*, p^*)$.
    \\

    \textbf{Inequality.} Let $\lim_{\sigma \to -\lambda_{min}^{+}} \phi(\sigma) > 0$, then $\lim_{\sigma \to -\lambda_{min}^{+}} \| p(\sigma)\| <\delta$. As mentioned before, $\sigma = -\lambda_{min}>0$ cannot satisfy the second optimality condition. In this case, so-called \textit{hard case}, we attempt to find a solution that lies on the boundary. For $\sigma^* = -\lambda_{min}$, this optimal solution is given by
        \begin{equation}\label{eq.ii}
            p^* = \hat{p}^* + z^*,
        \end{equation}
    where $\hat{p}^*  = -(B+\sigma^* I)^{\dagger}g$ is computed by \eqref{eq.hh} and $z^* =  \alpha u_{min}$. Vector $u_{min}$ is a unit eigenvector in the subspace associated with $\lambda_{min}$ and $\alpha$ is obtained so that $\|p^*\| = \delta$; i.e.,
        \begin{equation}\label{alpha/umin}
            \alpha = \sqrt{\delta^2 - \|\hat{p}^*\|^2}.
        \end{equation}
    The computation of $u_{min}$ depends on $\lambda_{min} = \min\{ \lambda_1, \gamma\}$. If $\lambda_{min} = \lambda_1$ then the first column of $P$ is a leftmost eigenvector of $B$, and thus, $u_{min}$ is set to the first column of $P_{\parallel}$. On other hand, if $\lambda_{min} = \gamma$, then any vector in the column space of $P_{\perp}$ will be an eigenvector of $B$ corresponding to $\lambda_{min}$. However, we avoid forming matrix $P_{\perp}$ to compute $P_{\perp}P_{\perp}^Tg$ in \eqref{eq.hh} if $\lambda_{min} = \lambda_1$. By the definition \eqref{eq.e}, we have
        \begin{equation*}
            \text{Range}(P_{\perp}) =\text{Range}(P_{\parallel})^{\perp},\qquad
            \text{Range}(P_{\parallel}) = \text{Ker}(I-P_{\parallel}P_{\parallel}^T).
        \end{equation*}
        To find a vector in the column space of $P_{\perp}$, we use $I-P_{\parallel}P_{\parallel}^T$ as projection matrix  mapping onto the column space of $P_{\perp}$. For simplicity, we can map one canonical basis vector at a time onto the column space of $P_{\perp}$ until a nonzero vector is obtained. This practical process, repeated at most $k+1$ times, will result in a vector that lies in $\text{Range}(P_{\perp})$; i.e.,
        \begin{equation}\label{umin}
            u_{min} \triangleq (I-P_{\parallel}P_{\parallel}^T)e_j,
        \end{equation}
    for $j = 1, 2, \dots k+1$ with $\|u_{min}\| \neq 0$; because $e_j \in \text{Range}(P_{\parallel})$ and
        $$\text{rank}(P_{\parallel}) = \text{dim}\,\text{Range}(P_{\parallel}) = \text{dim}\,\text{Kerl}(I-P_{\parallel}P_{\parallel}^T)=k.$$
\end{enumerate}
\autoref{alg:OBS} (see \autoref{app.Algs}) describes how to solve the TR subproblem for the optimal search direction $p^*$.

\section{Extended numerical results}\label{app.ExFig}
Further figures of numerical results on different classification problems listed in \autoref{Info} are provided in this section.

\begin{figure}[H]
\centering
\includegraphics[width=0.87\textwidth]{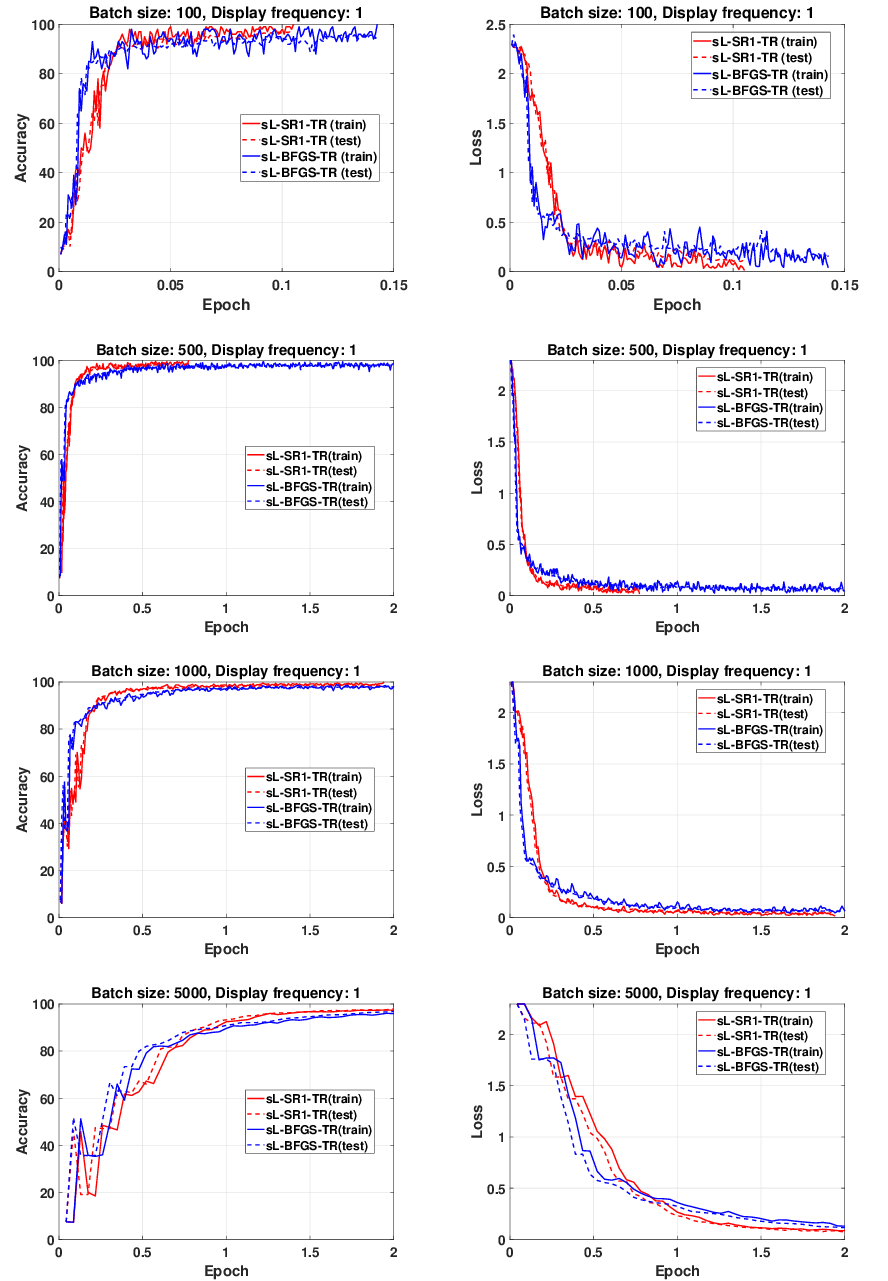}
\caption{MNIST with LeNet-like: Evolution of the training and testing loss and accuracy using stochastic training algorithms sL-BFGS-TR and sL-SR1-TR with $l=20$ and different batch sizes.}
\label{Mnist_Lenet}
\end{figure}


\begin{figure}[H]
\centering
\includegraphics[width=0.87\textwidth]{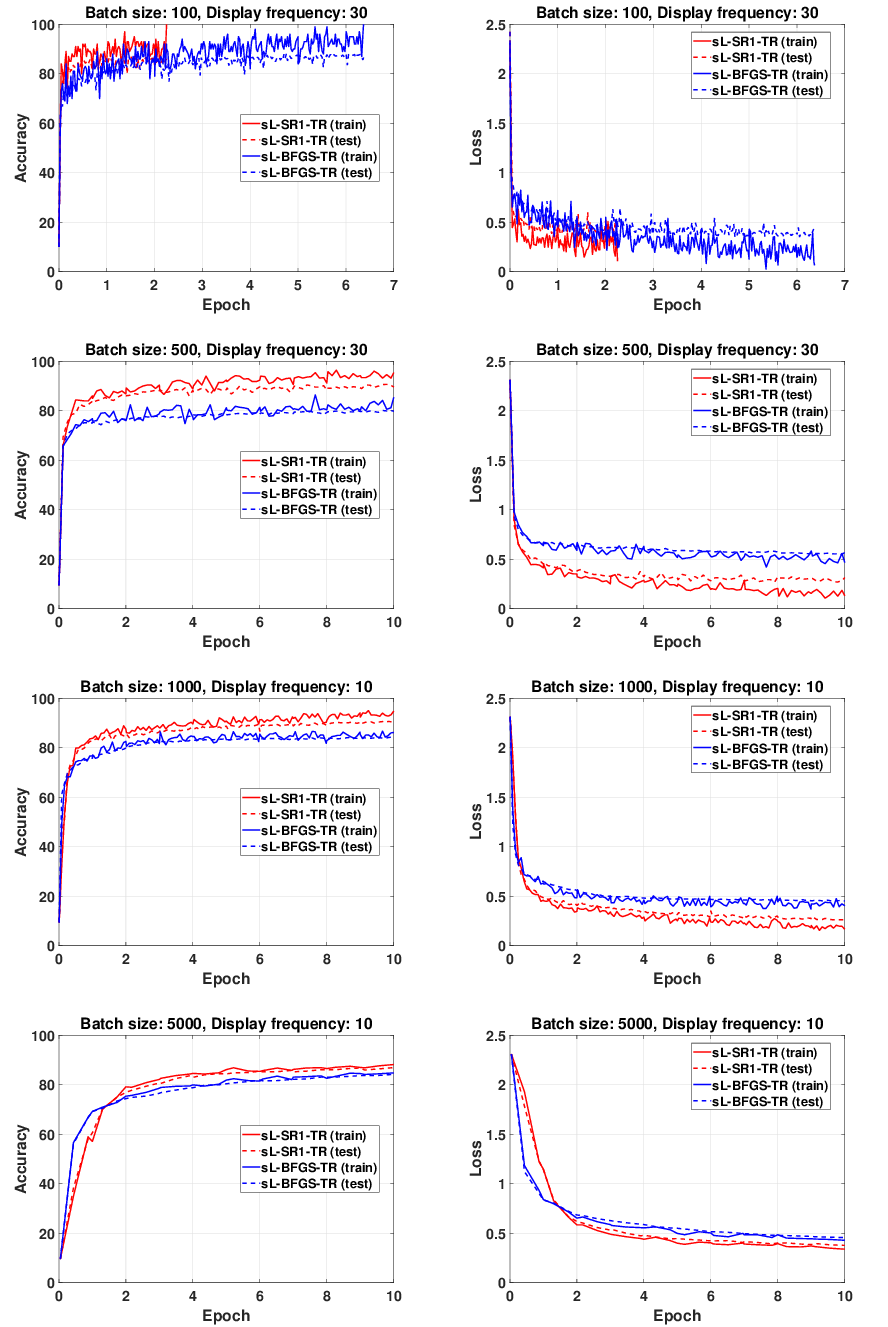}
\caption{Fashion-MNIST with LeNet-like: Evolution of the training and testing loss and accuracy using stochastic training algorithms sL-BFGS-TR and sL-SR1-TR with $l=20$ and different batch sizes.}
\label{F.Mnist_lenet}
\end{figure}


\begin{figure}[H]
\centering
\includegraphics[width=0.87\textwidth]{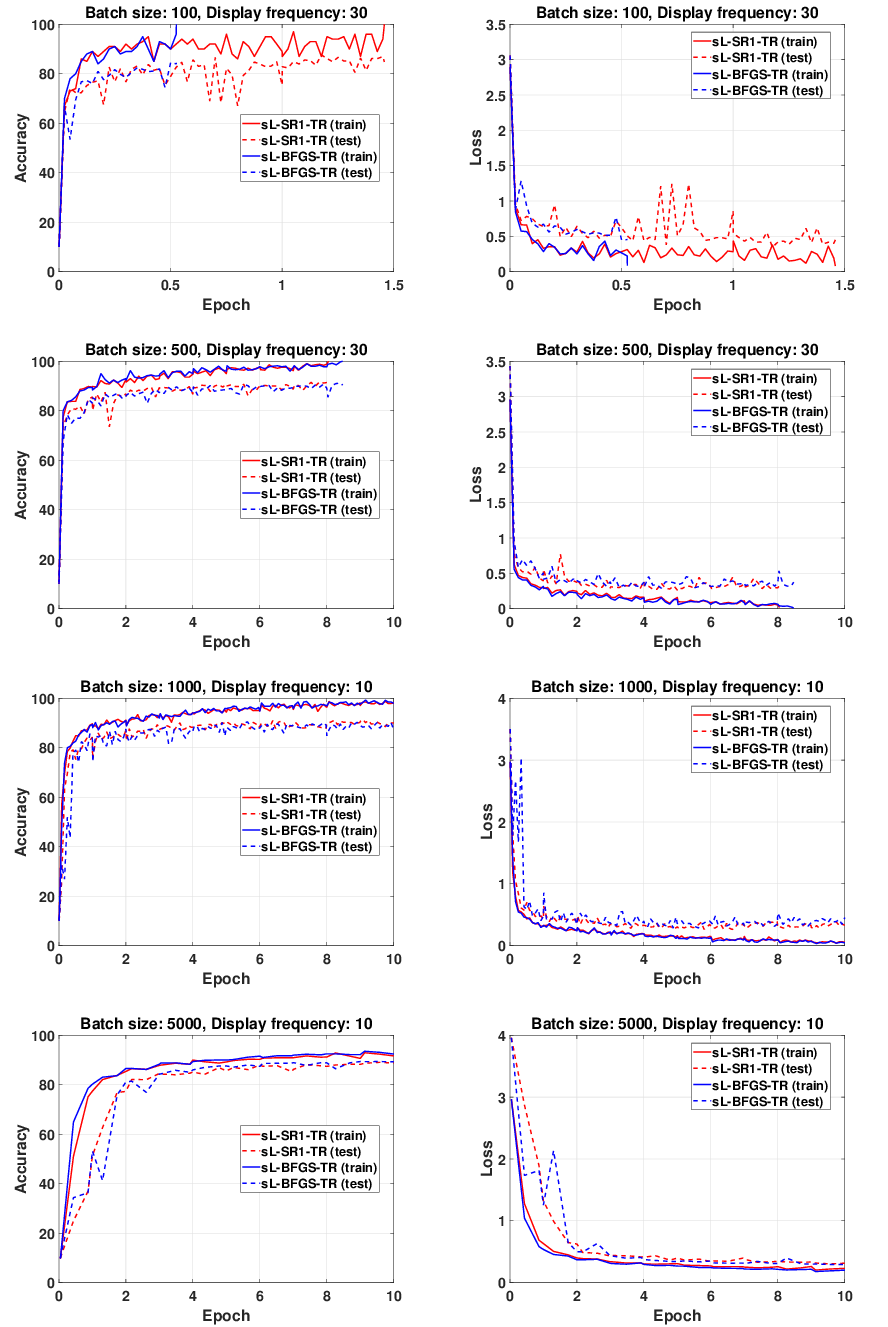}
\caption{Fashion-MNIST with ResNet-20: Evolution of the training and testing loss and accuracy using stochastic training algorithms sL-BFGS-TR and sL-SR1-TR with $l=20$ and different batch sizes.}
\label{F.Mnist_ResNet}
\end{figure}


\begin{figure}[H]
\centering
\includegraphics[width=0.87\textwidth]{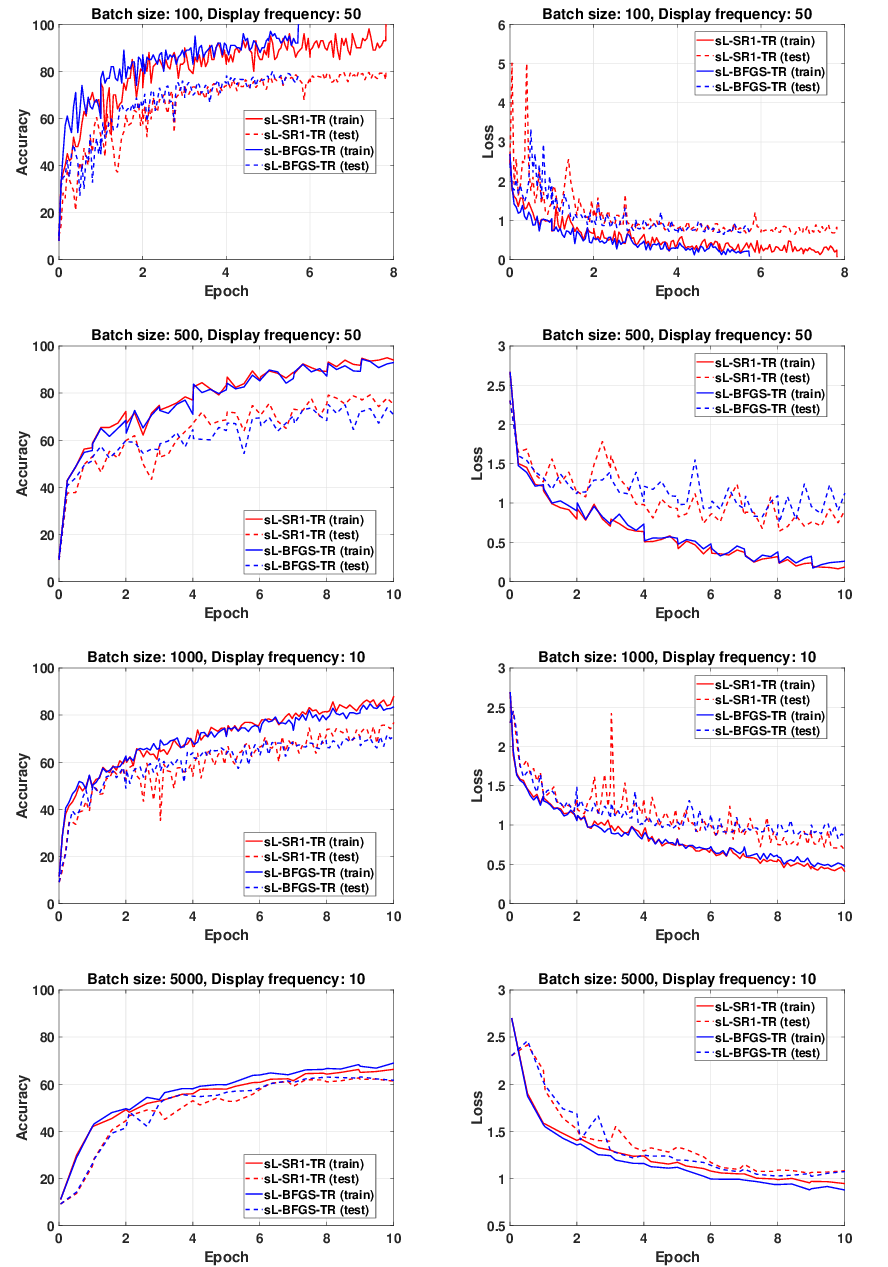}
\caption{CIFAR10 with ResNet-20: Evolution of the training and testing loss and accuracy using stochastic training algorithms sL-BFGS-TR and sL-SR1-TR with $l=20$ and different batch sizes.}
\label{Cifar10_ResNet}
\end{figure}


\begin{figure}[H]
\centering
\includegraphics[width=0.87\textwidth]{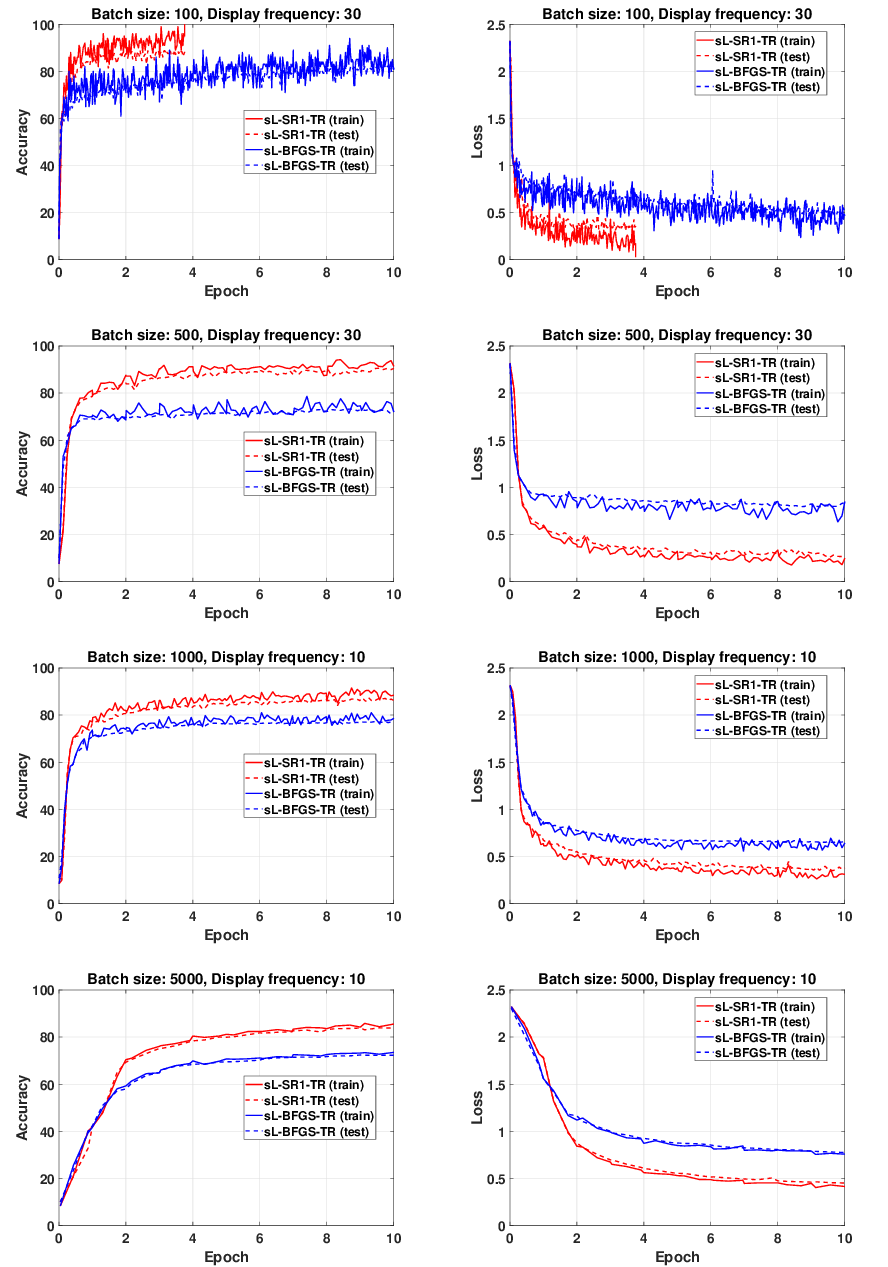}
\caption{Fashion-MNIST with ResNet-20(no BN): Evolution of the training and testing loss and accuracy using stochastic training algorithms sL-BFGS-TR and sL-SR1-TR with $l=20$ and different batch sizes.}
\label{FashionMnist_ResNet_noBN}
\end{figure}


\begin{figure}[H]
\centering
\includegraphics[width=0.87\textwidth]{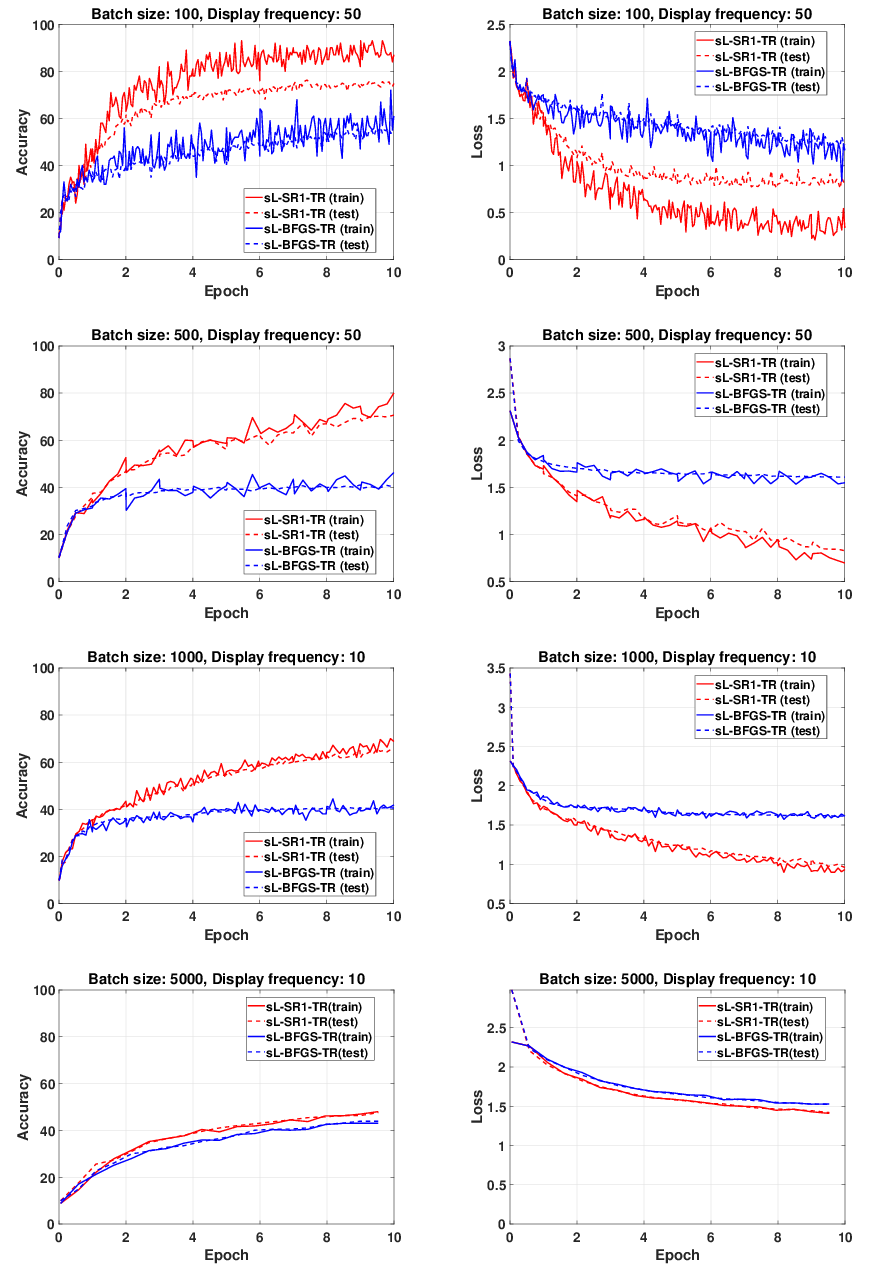}
\caption{CIFAR10 with ResNet-20(no BN): Evolution of the training and testing loss and accuracy using stochastic training algorithms sL-BFGS-TR and sL-SR1-TR with $l=20$ and different batch sizes.}
\label{Cifart_ResNet_noBN}
\end{figure}


\newpage
\begin{figure}[H]
\centering
\includegraphics[width=0.87\textwidth]{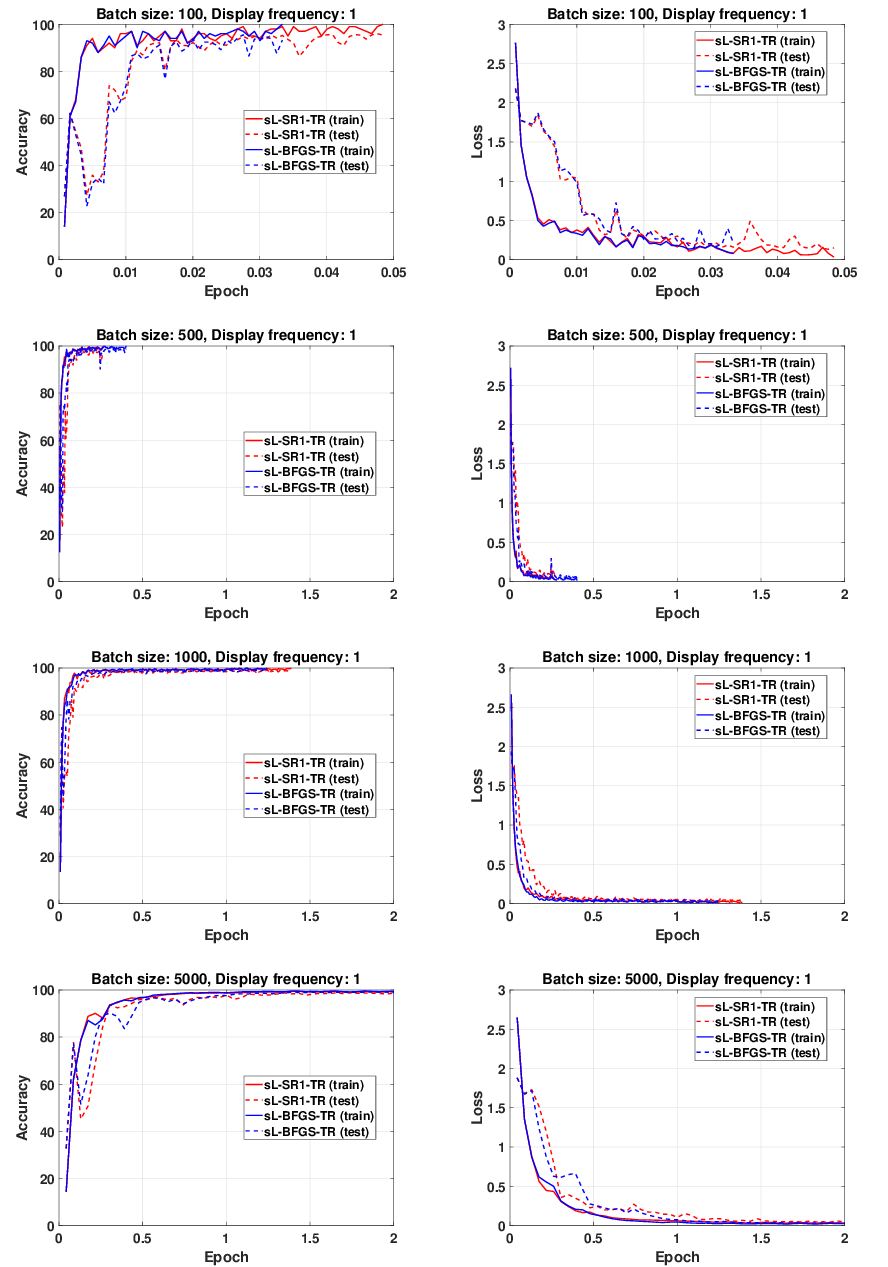}
\caption{MNIST with ConvNet3FC2: Evolution of the training and testing loss and accuracy using stochastic training algorithms sL-BFGS-TR and sL-SR1-TR with $l=20$ and different batch sizes.}
\label{MNIST_cnn(a)}
\end{figure}


\newpage
\begin{figure}[H]
\centering
\includegraphics[width=0.87\textwidth]{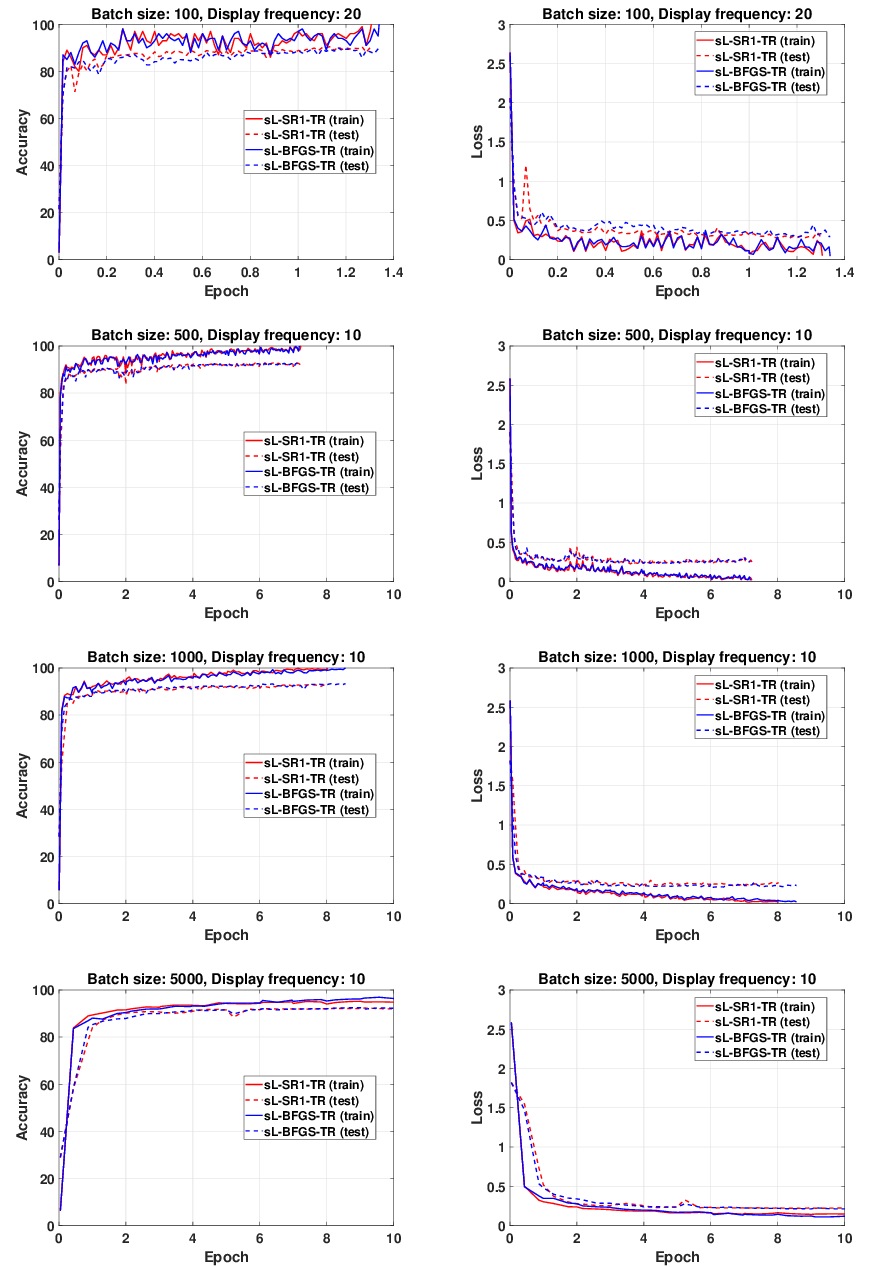}
\caption{Fashion-MNIST with ConvNet3FC2: Evolution of the training and testing loss and accuracy using stochastic training algorithms sL-BFGS-TR and sL-SR1-TR with $l=20$ and different batch sizes.}
\label{F.MNIST_cnn(a)}
\end{figure}


\newpage
\begin{figure}[H]
\centering
\includegraphics[width=0.87\textwidth]{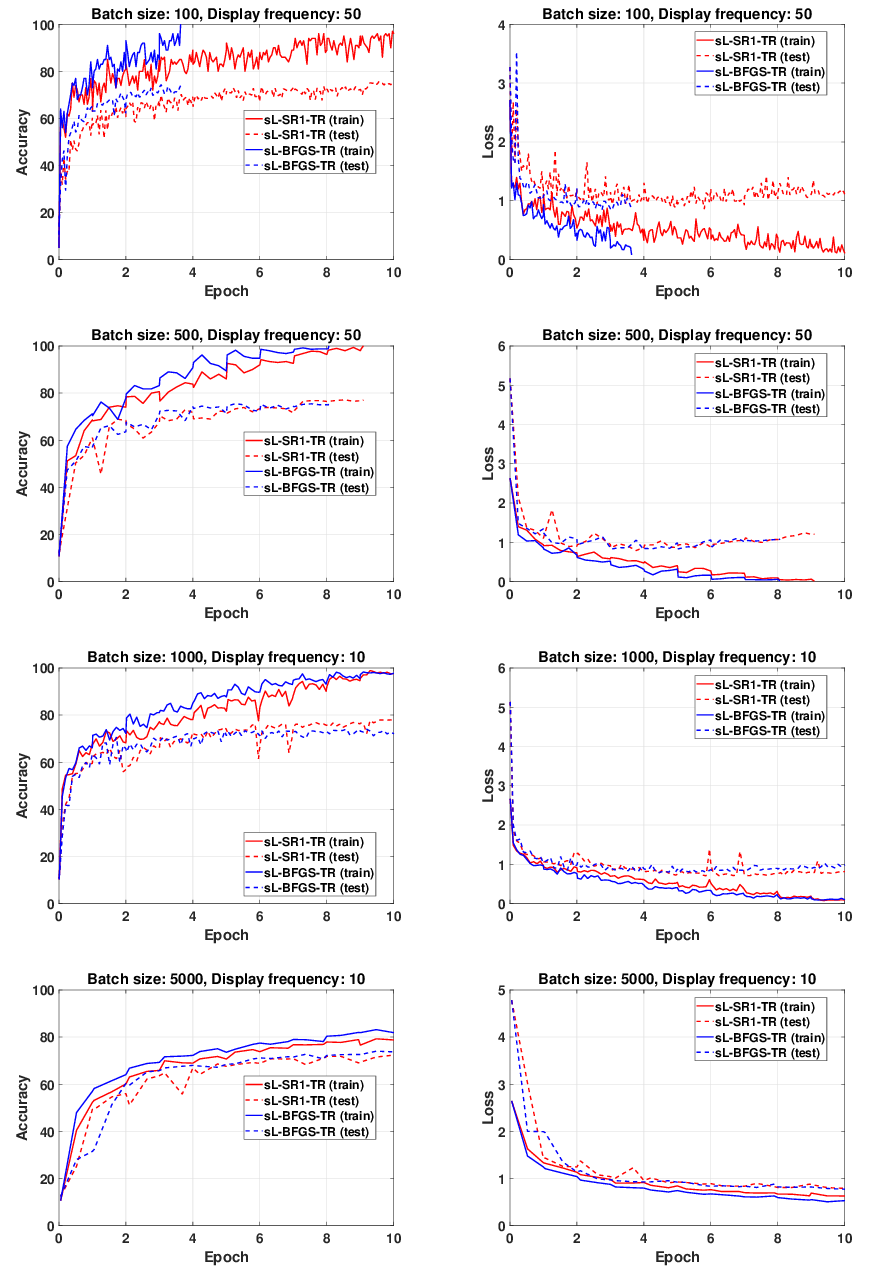}
\caption{CIFAR10 with ConvNet3FC2: Evolution of the training and testing loss and accuracy using stochastic training algorithms sL-BFGS-TR and sL-SR1-TR with $l=20$ and different batch sizes.}
\label{Cifar10_cnn(a)}
\end{figure}


\newpage
\begin{figure}[H]
\centering
\includegraphics[width=0.87\textwidth]{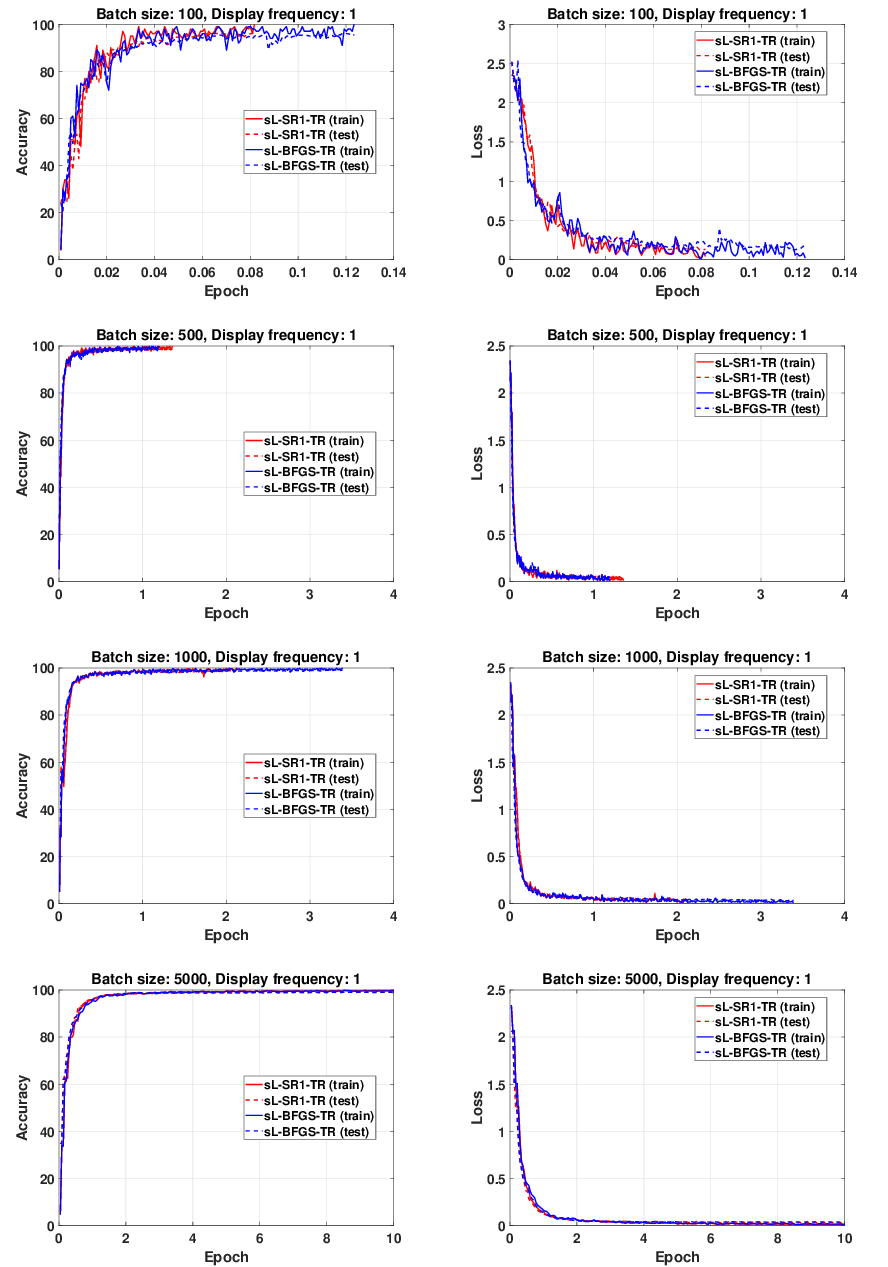}
\caption{MNIST with ConvNet3FC2(no BN): Evolution of the training and testing loss and accuracy using stochastic training algorithms sL-BFGS-TR and sL-SR1-TR with $l=20$ and different batch sizes.}
\label{MNIST_cnn(b)}
\end{figure}


\newpage
\begin{figure}[H]
\centering
\includegraphics[width=0.87\textwidth]{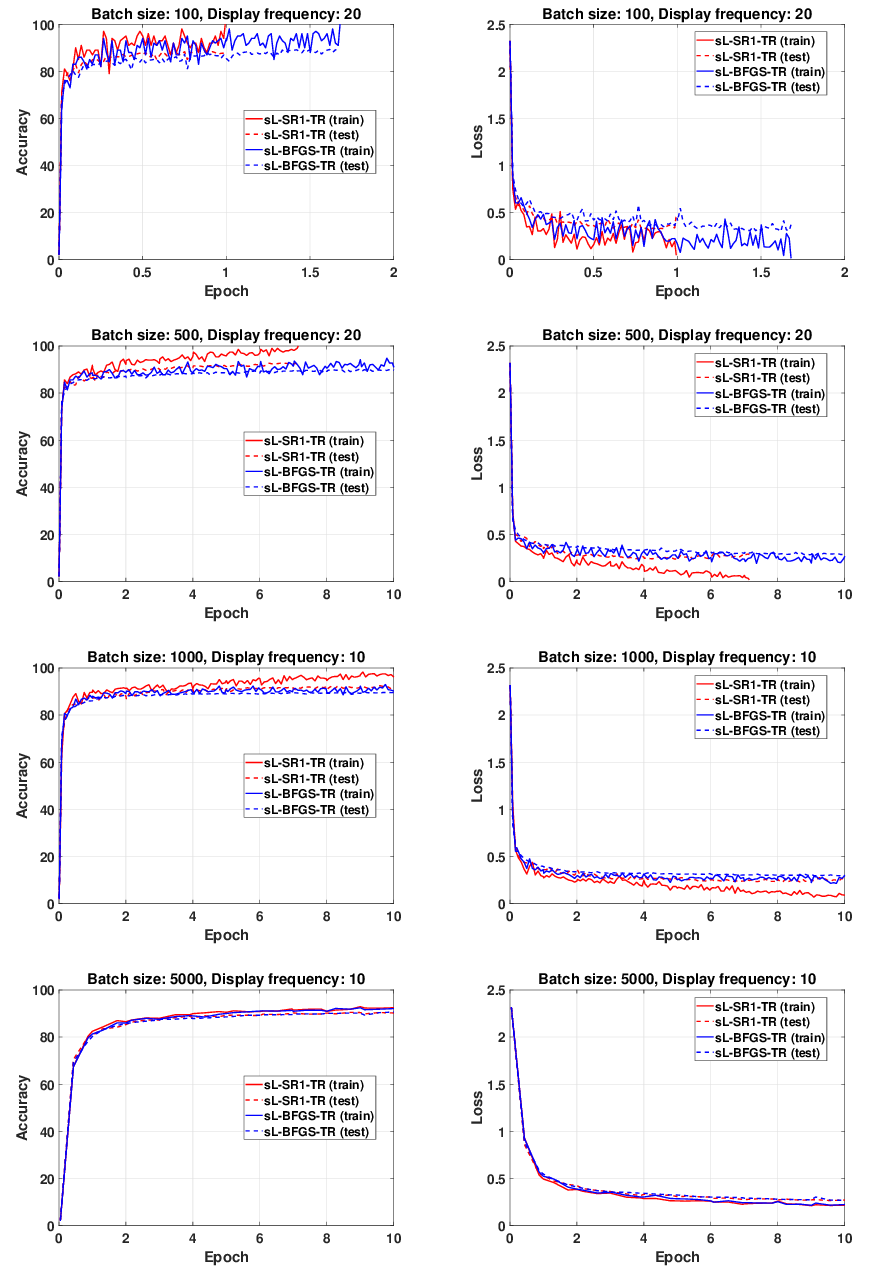}
\caption{Fashion-MNIST with ConvNet3FC2(no BN): Evolution of the training and testing loss and accuracy using stochastic training algorithms sL-BFGS-TR and sL-SR1-TR with $l=20$ and different batch sizes.}
\label{F.MNIST_cnn(b)}
\end{figure}


\newpage
\begin{figure}[H]
\centering
\includegraphics[width=0.87\textwidth]{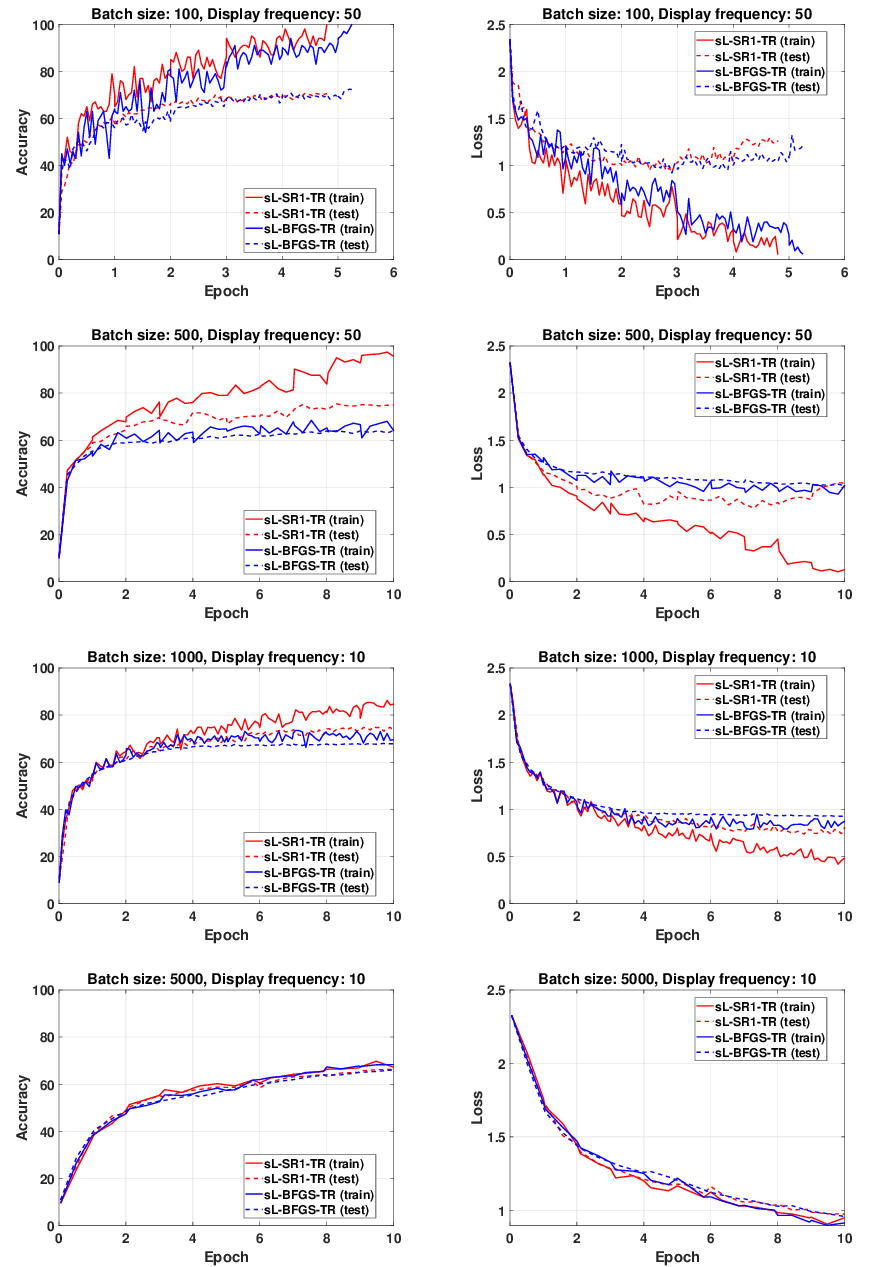}
\caption{CIFAR10 with ConvNet3FC2(no BN): Evolution of the training and testing loss and accuracy using stochastic training algorithms sL-BFGS-TR and sL-SR1-TR with $l=20$ and different batch sizes.}
\label{Cifar10_cnn(b)}
\end{figure}


\newpage
\begin{figure}[H]
\centering
\includegraphics[width=0.87\textwidth]{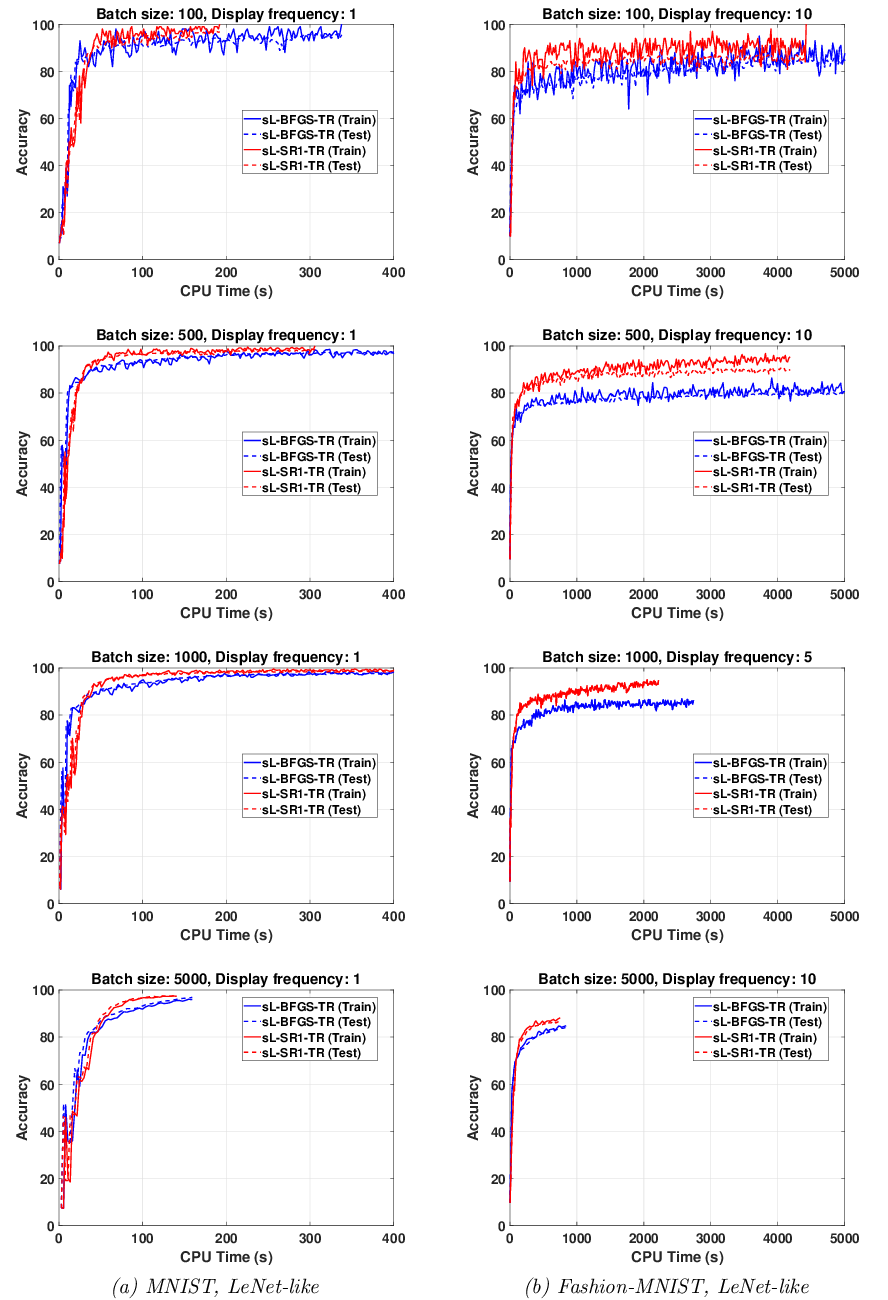}
\caption{Training CPU time (in seconds) of both algorithms with $l=20$.}
\label{fig_time_epoch1}
\end{figure}

\newpage
\begin{figure}[H]
\centering
\includegraphics[width=0.87\textwidth]{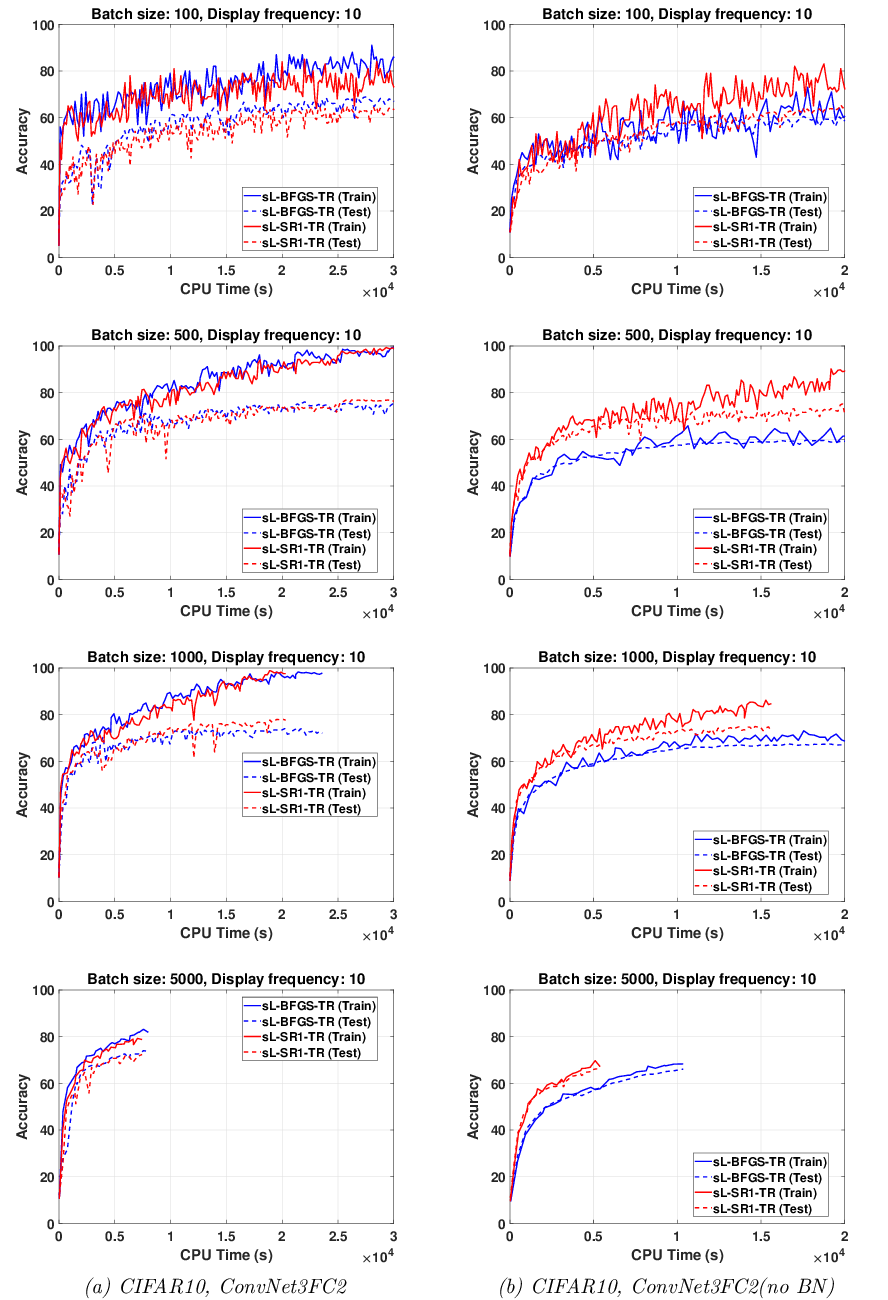}
\caption{Training CPU time (in seconds) of both algorithms with $l=20$.}
\label{fig_time_epoch3}
\end{figure}



\newpage
\begin{figure}[H]
\centering
\includegraphics[width=0.87\textwidth]{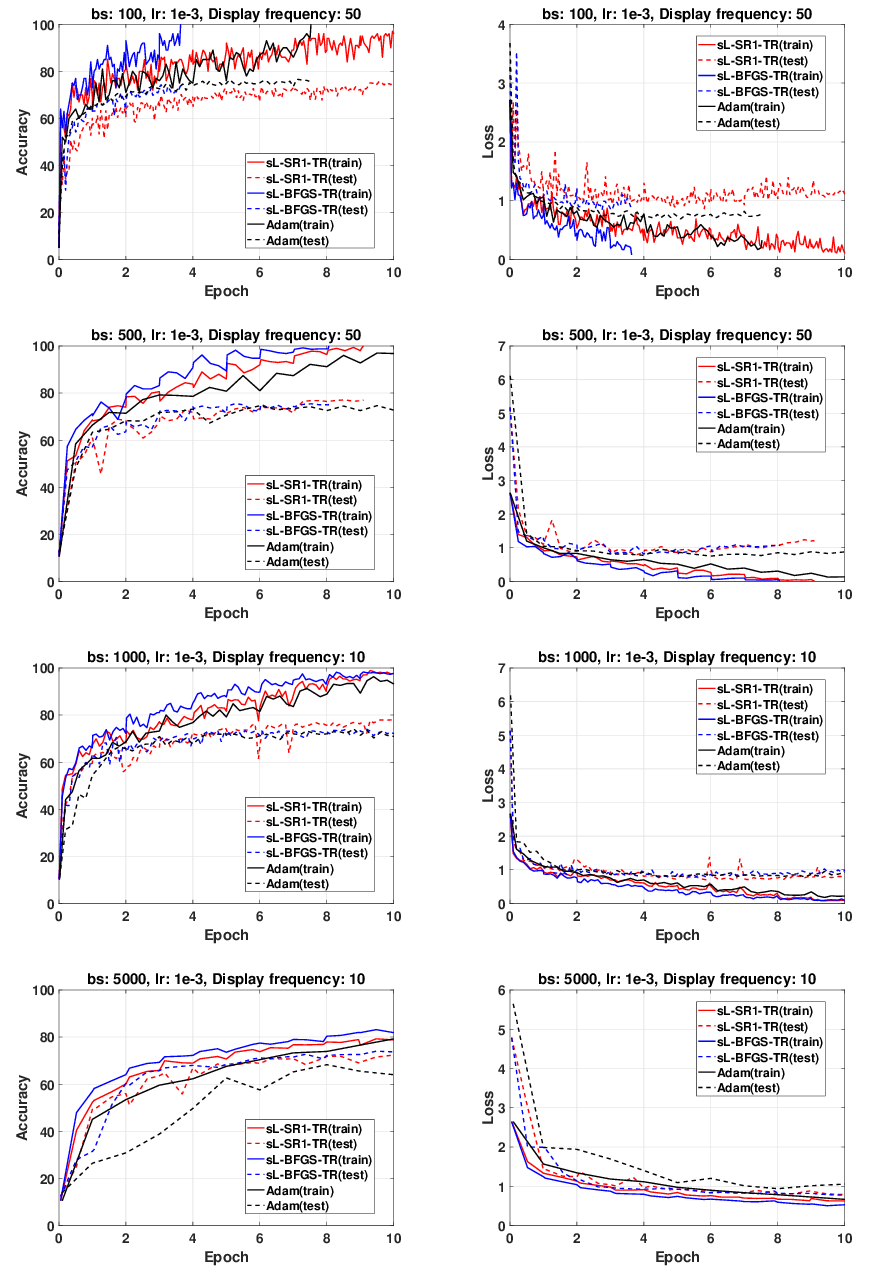}
\caption{CIFAR10 with ConvNet3FC2: Evolution of the training and testing loss and accuracy using sL-BFGS-TR, sL-SR1-TR with $l=20$ and \textit{tuned} Adam with optimal learning rate ($lr$) and different batch sizes ($bs$).}
\label{Cifar_Conv_a_adam}
\end{figure}

\newpage
\begin{figure}[H]
\centering
\includegraphics[width=0.87\textwidth]{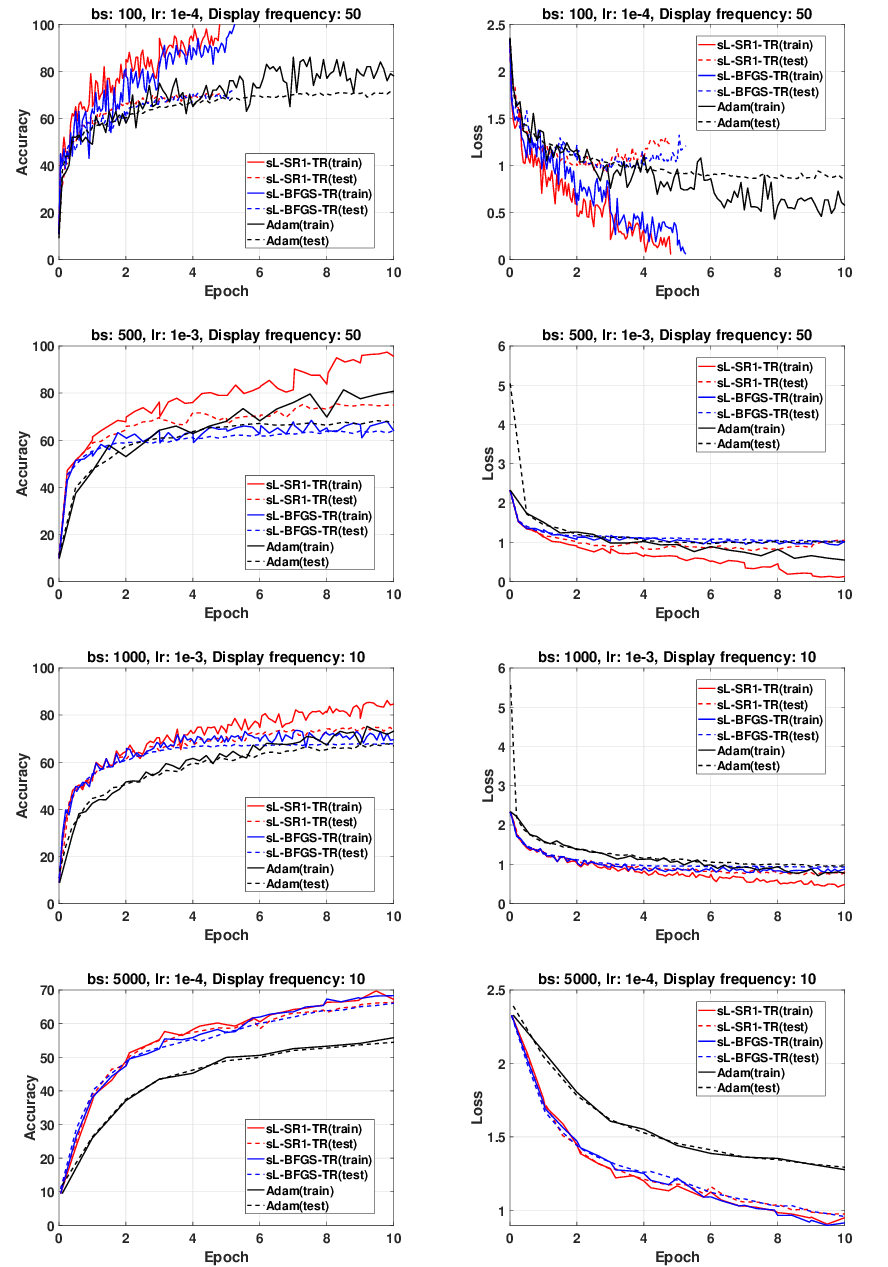}
\caption{CIFAR10 with ConvNet3FC2(no BN): Evolution of the training and testing loss and accuracy using sL-BFGS-TR, sL-SR1-TR with $l=20$ and \textit{tuned} Adam with optimal learning rate ($lr$) and different batch sizes ($bs$).}
\label{Cifar_Conv_b_adam}
\end{figure}

\newpage
\begin{figure}[H]
\centering
\includegraphics[width=0.87\textwidth]{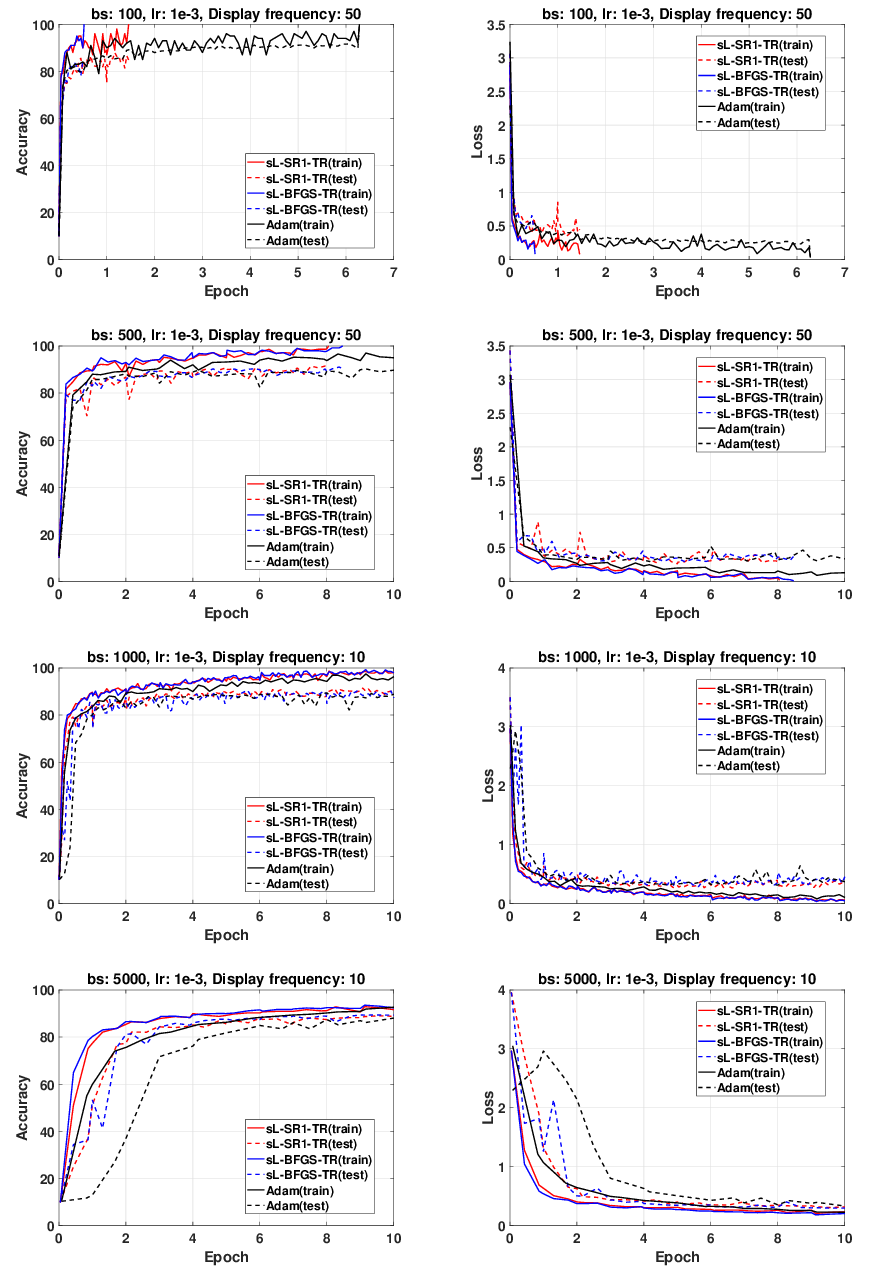}
\caption{Fashion-MNIST with ResNet-20: Evolution of the training and testing loss and accuracy using sL-BFGS-TR, sL-SR1-TR with $l=20$ and \textit{tuned} Adam with optimal learning rate ($lr$) and different batch sizes ($bs$).}
\label{Fmnist_ResNet_a_adam}
\end{figure}

\newpage
\begin{figure}[H]
\centering
\includegraphics[width=0.87\textwidth]{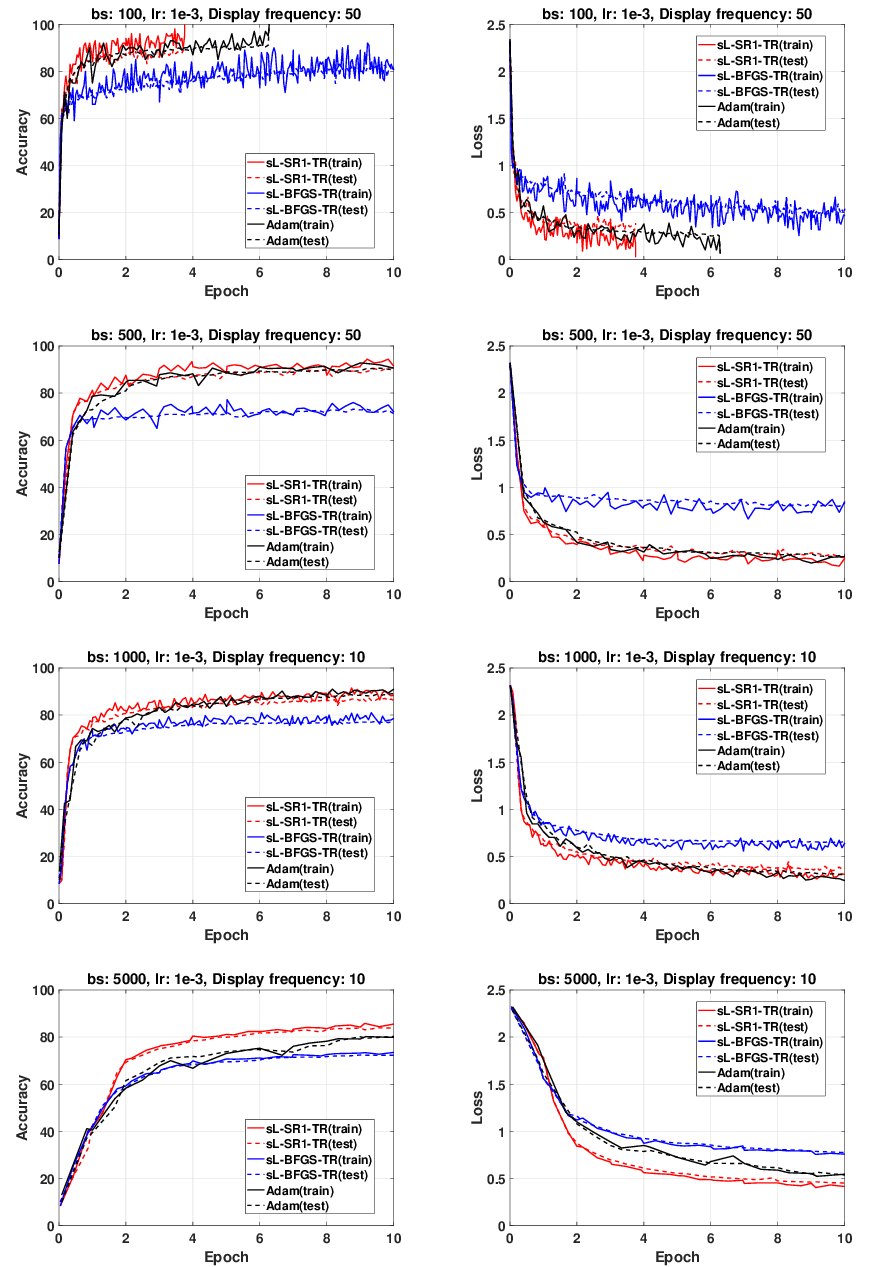}
\caption{Fashion-MNIST with ResNet-20(no BN): Evolution of the training and testing loss and accuracy using sL-BFGS-TR, sL-SR1-TR with $l=20$ and \textit{tuned} Adam with optimal learning rate ($lr$) and different batch sizes ($bs$).}
\label{Fmnist_ResNet_b_adam}
\end{figure}
\begin{figure}[H]
\centering
\includegraphics[width=0.87\textwidth]{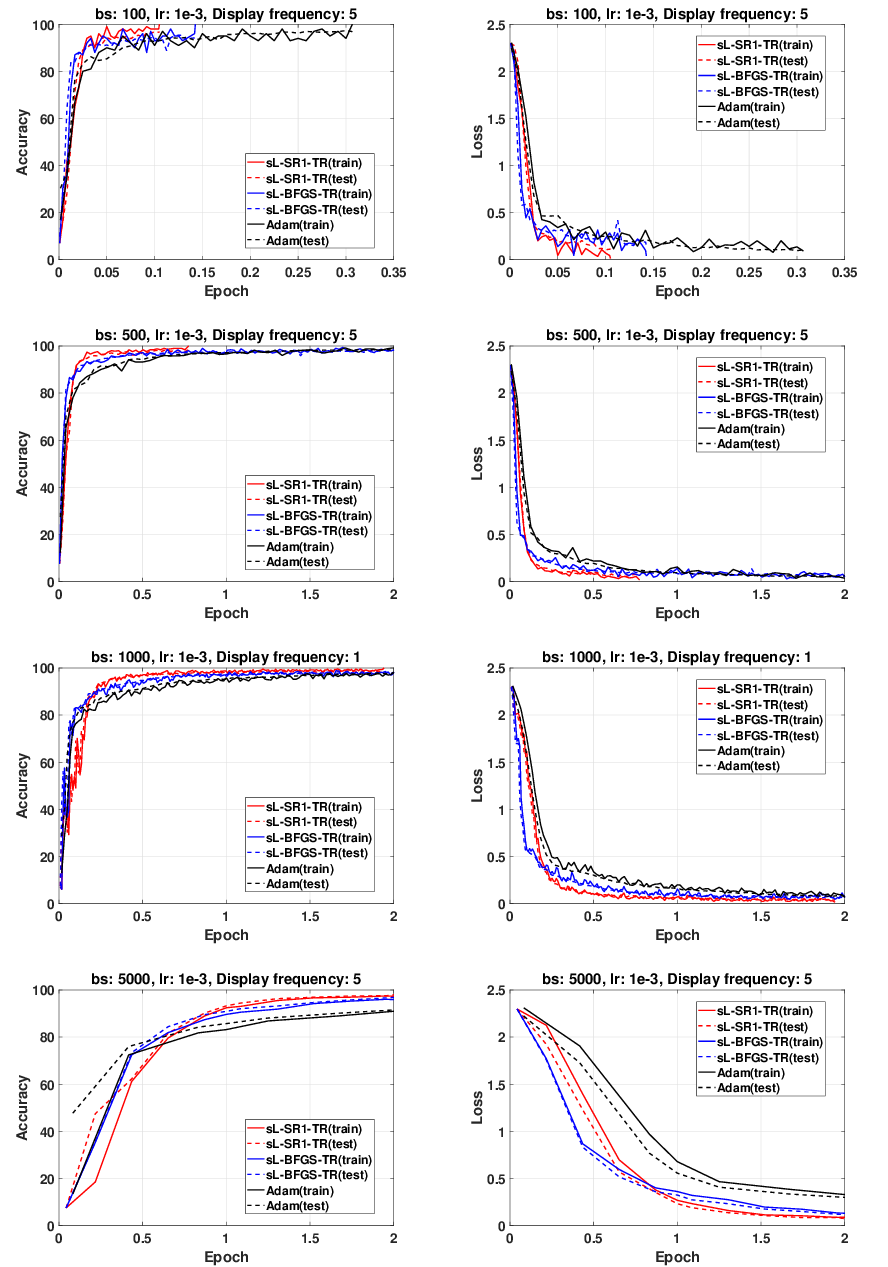}
\caption{MNIST with LeNet-like: Evolution of the training and testing loss and accuracy using sL-BFGS-TR, sL-SR1-TR with $l=20$ and \textit{tuned} Adam with optimal learning rate ($lr$) and different batch sizes ($bs$).}
\label{Mnist_LeNet_adam}
\end{figure}

\end{document}